\documentclass{article}

\usepackage{arxiv}
\usepackage{cite}
\usepackage[utf8]{inputenc} 
\usepackage[T1]{fontenc}    
\usepackage{hyperref}       
\usepackage{url}  
\usepackage{float}
\usepackage{booktabs}       
\usepackage{amsfonts}       
\usepackage{nicefrac}       
\usepackage{microtype}      
\usepackage{lipsum}		
\usepackage{graphicx}
\usepackage{doi}

\title{Learnable Triangulation for Deep Learning-based 3D Reconstruction of Objects of Arbitrary Topology from Single RGB Images}

\author{ Tarek~Ben~Charrada~\thanks{Tarek Ben Charrada is with  the Informatique, BioInformatique, Systèmes Complexes EA 4526 laboratory, Univ Evry, Université Paris-Saclay, Evry-Courcouronnes, 91000, France and with Ausy, Sèvres,  92130, France.} \\
	\texttt{tarek.ben-charrada@ensea.fr} \\
    
	\And
	Hedi~Tabia~\thanks{Hedi Tabia is with  the Informatique, BioInformatique, Systèmes Complexes EA 4526 laboratory, Univ Evry, Université Paris-Saclay, Evry-Courcouronnes, 91000, France.}\\
	\texttt{hedi.tabia@univ-evry.fr} \\
	
	\And
	Aladine~Chetouani~\thanks{Aladine Chetouani is with Pluridisciplinaire de Recherche en Ingénierie des Systèmes, Mécanique et Energétique laboratory, Université Orléans, Orléans, 45100,, France.}\\
	\texttt{aladine.chetouani@univ-orleans.fr}
	\And
	Hamid~Laga~\thanks{Hamid Laga is with the Information Technology, Mathematics and Statistics Discipline, Murdoch University (Australia), and with the Phenomics and Bioinformatics Research Centre, University of South Australia.}\\
	\texttt{H.Laga@murdoch.edu.au}\\
}

\date{}

\renewcommand{\shorttitle}{\phantom{learnable triangulation}}

\hypersetup{
pdftitle={Learnable Triangulation for Deep Learning-based 3D Reconstruction of Objects of Arbitrary Topology from Single RGB Images},
pdfsubject={q-bio.NC, q-bio.QM},
pdfauthor={Tarek~Ben~Charrada Hedi~Tabia Aladine~Chetouani Hamid~Laga},
pdfkeywords={3D reconstruction, Arbitrary topology,Mesh generation},
}
\usepackage{times}
\usepackage{epsfig}
\usepackage{graphicx}
\usepackage{amsmath}
\usepackage{amssymb}

\usepackage{xcolor}
\usepackage{microtype}
\usepackage{cite}
\usepackage{float}
\usepackage{multirow}
\usepackage{capt-of}
\usepackage{breakcites}
\usepackage{booktabs}
\usepackage{amsmath}
\usepackage{algorithm,algpseudocode}
\usepackage{caption}
\usepackage{subcaption}

\newcommand{\period}{T}			
\newcommand{\explorability}{\epsilon}
\newcommand{\logcmd}{\text{log}_{\text{CMD}}}
\newcommand{\statespace}{\mathcal{S}}
\newcommand{\coefcmd}{\lambda_1}

\newcommand{\coeflc}{\lambda_{3}}
\newcommand{\coefel}{\lambda_{4}}
\newcommand{\coefnc}{\lambda_{5}}
\newcommand{\coefnl}{\lambda_{6}}
 
 \newcommand{\coefogcmd}{\lambda_{2}}

\newcommand{\Ncgn}{\mathcal{N}_{CGN}}

\newcommand{\ie}{\emph{i.e., }}
\newcommand{\eg}{\emph{e.g., }}
\newcommand{\etal}{\emph{et al.}}

\newcommand{\noi}{\noindent}

\usepackage{hyperref}
\begin{document}
\maketitle

\begin{abstract}
We propose a novel deep reinforcement learning-based  approach for 3D object reconstruction from monocular images. Prior works that use mesh representations are template based. Thus, they are limited to the reconstruction of objects that have the same topology as the template. Methods that use volumetric grids as intermediate representations are computationally expensive, which limits their application in real-time scenarios. In this paper, we propose a novel end-to-end method that reconstructs 3D objects of arbitrary topology from a monocular image. It is composed of of (1) a Vertex Generation Network (VGN), which predicts the initial 3D locations of the object’s vertices from an input RGB image, (2) a differentiable triangulation layer, which learns in a non-supervised manner, using a novel reinforcement learning algorithm, the best triangulation of the object's vertices,  and finally, (3) a hierarchical mesh refinement network that uses graph convolutions to refine the initial mesh. Our key contribution is the learnable triangulation process, which recovers in an unsupervised manner the topology of the input shape. Our experiments on  ShapeNet and Pix3D benchmarks show that the proposed method outperforms the state-of-the-art in terms of visual quality, reconstruction accuracy, and computational time.

\end{abstract}


\section{Introduction}
\label{sec:introduction}

3D shape reconstruction from one or multiple images is a long-standing ill-posed problem that has been extensively investigated by the computer vision, graphics, and machine learning communities. Despite the large body of research, image-based 3D reconstruction remains a challenging problem.  With the  success of deep learning techniques in many vision tasks, several recent papers have addressed the problem using Convolutional Neural Networks, see~\cite{han2019image,laga2020survey} for a comprehensive survey. In these works, 3D reconstruction is formulated as an inference problem taking advantage of the availability of collections of images annotated with their corresponding 3D models~\cite{mehta2017vnect,luvizon2020multi}.

Existing deep learning methods for 3D reconstruction represent 3D models as volumes~\cite{choy20163d,wu2017marrnet,tulsiani2017multi,sun2018pix3d}, point clouds~\cite{fan2017point,lin2018learning,Jiang_2018_ECCV,mandikal20183d,gadelha2018multiresolution,li2018point,sun2018pointgrow,li2018efficient,li2018optimizable,zeng2018inferring,mandikal2019dense,wang2018mvpnet}, or triangulated meshes~\cite{wang2018pixel2mesh,kato2018neural,kanazawa2018learning,pan2019deep}. Volumetric representations are suitable for convolutional operations, which operate on regular grids. Although they can reconstruct 3D shapes of arbitrary topologies, they are very expensive in terms of memory requirements, and thus are limited to the reconstruction of low resolution 3D models. Point-based methods can efficiently represent complex shapes, but rely on post-processing to extract the meshes. Techniques that use mesh representations learn how to deform pre-defined templates to fit the input~\cite{wang2018pixel2mesh,kato2018neural,kanazawa2018learning}. As a result, the class of objects that can be reconstructed with these methods is limited to those that have the same topology as the template. The recent attempts to overcome this limitation, \eg~\cite{gkioxari2019mesh}, use a volumetric grid as an intermediate representation in order to recover an initial 3D geometry of the same topology as the input, followed by a Graph Convolutional Network (GCN) to infer the triangulated mesh from the volumetric grid. Thus, these methods inherit the same limitations as the volumetric methods.

We propose in this paper a novel algorithm that is able to reconstruct, from a single RGB image,  3D objects of arbitrary genus in the form of triangulated meshes. Unlike previous mesh-based representations, which rely on template deformation or volumetric grids as an intermediate representation, the proposed approach learns to infer both the topology and the geometry of the 3D objects. Our approach is composed of a Vertex Generation Network (VGN), which predicts the initial 3D locations of the  object's vertices from an input RGB image, a differentiable triangulation layer whose parameters  are automatically learned using a novel reinforcement learning-based algorithm, and a hierarchical mesh refinement network that uses graph convolutions to refine the initial mesh.  Our main  contributions are:
\begin{itemize}
    \item A novel representation and a novel deep network that enable mesh-based 3D reconstruction of objects of arbitrary topology without relying on template deformations or volumetric representations.

   \item  A novel learnable triangulation layer, based on reinforcement learning, which enables the transition from a point cloud  to a mesh-based surface representation. The latter enables high resolution reconstruction of surfaces of arbitrary genus.

    \item A novel loss function based on the log Chamfer distance to overcome, with a minimum computational overhead, the limitations of the traditional Chamfer distance-based losses. 
    \item We show that traditional  accuracy measures do not correlate with the visual plausibility and thus novel metrics are required to assess the performance of 3D reconstruction algorithms. 

\end{itemize}

\noindent The proposed framework is lightweight and can reconstruct, with high accuracy,  high resolution 3D models of varying topological structures without relying on volumetric representations or template deformation. The remainder of the paper is organised as follows; Section~\ref{sec:related_work} reviews the related work. Section~\ref{sec:method} describes in detail the proposed method. Section~\ref{sec:results} presents the results and evaluates the proposed approach in terms of accuracy, performance, and visual plausibility. Section~\ref{sec:conclusion} summarizes the main findings of the paper.

\begin{figure*}[t]

  \begin{center}
    \includegraphics[width=\linewidth,trim={1.3cm 10.4cm 11.1cm 1cm},clip]{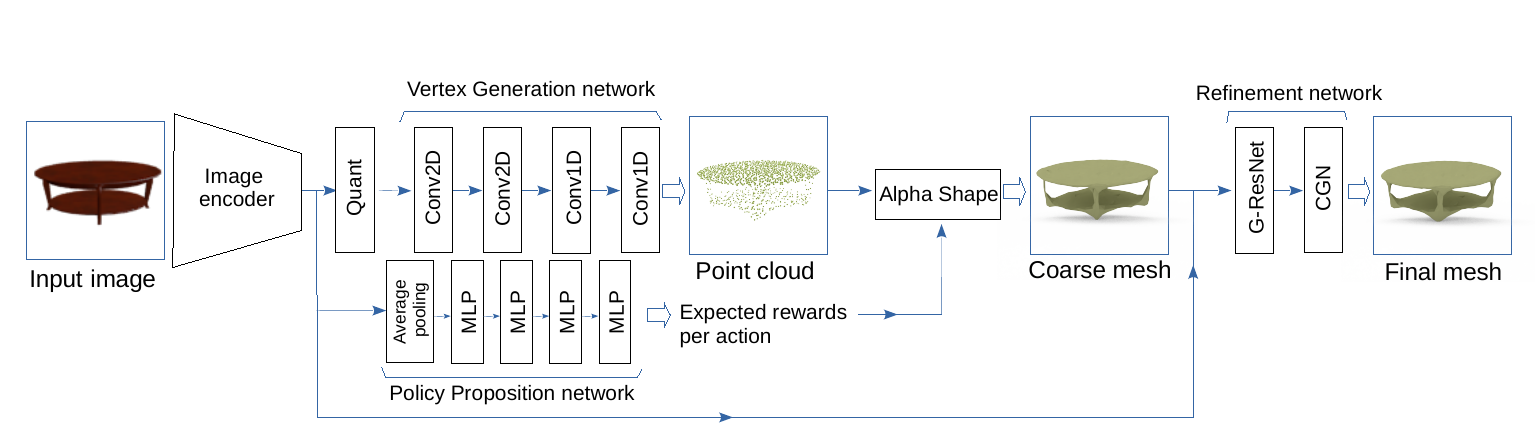}
    \end{center}
\caption{The proposed pipeline is composed of \textbf{(1)} a feature extractor, which maps the input RGB image into a latent representation, \textbf{(2)} a Vertex Generation Network (VGN), which generates, from the latent representation, an initial point cloud to represent the initial coordinates of the shape vertices, \textbf{(3)} a Policy Proposition Network (PPN), which triangulates the initial vertices and generates a mesh that captures the topology of the object, and \textbf{(4)} a Mesh Refinement Network (MRN), which hierarchically refines the coarse geometry using a Graph Convolutional Network (GCN).}
\label{fig:overview}
\end{figure*}

\section{Related Work}
\label{sec:related_work}
Image-based 3D reconstruction   has been extensively investigated in the literature. Early methods  such as stereo matching~\cite{scharstein2002taxonomy}, visual hull reconstruction, and Structure-from-Motion (SfM)~\cite{schonberger2016structure}, are mainly based on multi-view geometry~\cite{hartley2003multiple}.  In recent years, however, we have seen a growing interest in formulating 3D reconstruction from RGB images as a learning task. In particular, with the success of deep learning networks and the availability of large annotated datasets, these methods start to achieve great success. Han \etal~\cite{han2019image} provide a comprehensive survey of the recent developments in this field. In this section, we classify the state-of-the-art based on the shape representations they use and discuss their advantages and limitations based on their ability to operate in real time and their ability to reconstruct 3D shapes of arbitrary topologies.

Early methods use volumetric representations such as occupancy maps~\cite{liu2017learning,Tulsiani_2018_CVPR},  (truncated) Signed Distance Functions (SDF)~\cite{dai2017shape,Cherabier_2018_ECCV,kundu20183d,Cao_2018_ECCV}, or  Deep SDF (deepSDF)~\cite{park2019deepsdf}. Their main advantage  is that many of the existing deep learning architectures that have been designed for image analysis can easily be extended to 3D data by replacing the 2D pixel array with its 3D analogue and then processing the grid using 3D convolution and pooling operations. They also allow for the reconstruction of surfaces of arbitrary topologies. They are, however, very expensive in terms of memory requirements and computation time. Thus, they are limited to low resolution reconstructions and are not suited for low memory devices. Although several techniques that exploit the sparsity of the volumetric representations have been introduced, \eg Octree Networks~\cite{riegler2017octnet}, Octree Generative Networks (OGN)~\cite{tatarchenko2017octree}, and Hierarchical Space Partitioning~\cite{hane2019hierarchical}, they are still limited due to the complexity of their implementations.  

3D shapes can also be represented as non-ordered sets of points, providing simplicity and efficiency in terms of memory requirements. Such point-based representations are well suited for objects with intriguing parts and fine details.  As such, an increasing number of papers, \eg \cite{fan2017point,lin2018learning,Jiang_2018_ECCV,mandikal20183d,gadelha2018multiresolution,li2018point,sun2018pointgrow,insafutdinov2018unsupervised,li2018efficient,li2018optimizable,zeng2018inferring,wang2018mvpnet,mandikal2019dense},   explored their usage for deep learning-based reconstruction. To reconstruct point clouds from an input image, these methods also use an encoder, similar to volumetric representations. They, however, differ in how the latent variable is decoded. Methods that represent point clouds as a matrix of size $N\times 3$, where $N$ is the number of points,  use fully connected layers~\cite{fan2017point,mandikal20183d,gadelha2018multiresolution,li2018point,insafutdinov2018unsupervised,mandikal2019dense}. Methods that use grid representations~\cite{fan2017point,wang2018mvpnet,lin2018learning,tatarchenko2016multi,li2018efficient} decode the latent representation using 2D  up-convolutional layers. While being memory efficient, these methods require setting in advance the maximum number of points. They also lack the notion of surface and thus require a post-processing step to generate the triangular meshes for visualization.

Methods that use mesh representations are template-based. The template can be either pre-defined, \eg  a sphere~\cite{kato2018neural,pan2019deep} or an ellipsoid~\cite{wang2018pixel2mesh}, retrieved from a database of exemplars~\cite{kurenkov2017deformnet,pontes2017image2mesh,jack2018learning}, or learned from data using a statistical model~\cite{tulsiani2017learning}.  A neural network composed of an encoder and a top network, \eg a  Graph Convolutional  Network (GCN) as in Pixel2Mesh of \cite{wang2018pixel2mesh}, then learns  a vector field that deforms the template to match the input.  The main limitation of these methods is that the  objects that can be reconstructed should be of the same topology as the template. Mesh R-CNN~\cite{gkioxari2019mesh} attempted to address this issue by using a voxel grid as an intermediate representation, combined with a GCN to refine the reconstruction. As such, it inherits the same limitations as the volumetric representations.

The approach we propose in this paper is mesh-based. Unlike  previous techniques, it is lightweight and can reconstruct 3D models of arbitrary topology with high accuracy and without relying on volumetric representations or template deformation.


\section{Method}
\label{sec:method}
In this paper, we focus on the high visual quality reconstruction of 3D objects of arbitrary topology from a single RGB image. We propose, for the first time,  a two-stage algorithm that is able to reconstruct 3D meshes of arbitrary topology without relying on template deformation or voxel representations, see Figure~\ref{fig:overview}. Using a novel Vertex Generation Network (VGN), the approach first generates an initial estimate of the object's 3D geometry by inferring the 3D locations of the object's vertices. These are then further processed with a learnable triangulation layer to generate an initial triangulation that captures   the topology of the target 3D shape, and thus enables the reconstruction of objects of arbitrary topology without relying on volumetric representations. In the second stage, the reconstructed coarse geometry is hierarchically refined using a Mesh Refinement Network (MRN) implemented using a Graph Convolutional Network (GCN). The entire pipeline is trained in an end-to-end fashion using a novel log Chamfer distance measured using points sampled from the reconstructed and ground truth meshes. 
Figure~\ref{fig:overview} provides an overview of the pipeline. Below, we discuss these components in details.

\subsection{Coarse Mesh Generation Block }
\label{sec:coarse_generation}

We infer a coarse 3D geometry of the object in the input image using a Vertex Generation Network (VGN), which returns the initial locations of the object vertices in the form of a point cloud (Section~\ref{sec:geometry_generation}). Directly triangulating this  point cloud can result in a very noisy triangular mesh. Also, existing triangulation algorithms are not learnable and thus are not suitable for training deep networks in an end-to-end fashion. To overcome these two limitations, we propose a novel learnable alpha shape triangulation block, which will be implemented using a novel Policy Proposition Network (PPN)  followed by a novel triangulation layer, which  recovers the topology of the  object observed in the input image (Section~\ref{sec:policy_prediction}).

\subsubsection{Vertex Generation Network (VGN)}
\label{sec:geometry_generation}

The VGN takes a latent representation generated by an image encoder and infers an initial coarse geometry in the form of 3D vertices. As shown in Figure~\ref{fig:overview}, a ResNet 50~\cite{he2016deep} encoder composed of a series of 2D convolutional layers  maps an input RGB image of size $137\times137$  to a latent representation of dimension  $5\times 5\times2048$. The latent representation is then discretized using vector quantisation~\cite{oord2017neural}  to  help prevent the network from overfitting by removing the decimal parts of the latent representation which often encode object specific details. Also, quantisation helps generate consistent reconstructions as the subsequent layers would not be affected by the noise often encoded in the decimal part. In fact, removing the decimal part of the latent representation is similar to dropout  regularization where some of the information learnt during the training stage is dropped to enhance the generalization capability of the deep learning model. Additionally, a discrete representation is more robust as it is less sensitive to fine grained details that might be instance-specific details. This robust representation favors generalization capability over accurate reconstructions.  The discrete latent representation is then decoded with a series of two consecutive CNNs, with output channels of size $1024$ and $512$, and kernels of size $3\times3$ and $2\times 2$, respectively. We append at the end  two transposed 1D deconvolution layers (Deconv1D) with output channels of size $512$ and  $1536$, respectively. Both layers have a kernel of size $3\times3$ and a padding of $1$. The first Deconv1D layer uses a stride of $2$. 
This produces an initial reconstruction of a coarse 3D geometry.

\subsubsection{Learnable Alpha Shape Triangulation}
\label{sec:policy_prediction}
There are several algorithms that one can use to triangulate the point cloud generated by the vertex generation network. One of the popular ones is the alpha shape algorithm~\cite{edelsbrunner1983shape}, which operates in two stages. It first generates an initial tetrahedral faces using Delaunay complex and then filters out the large ones by removing the tetrahedrals with a circumradius that is larger than a predefined threshold $\tau$. We refer to this last stage as the filtering stage. While the algorithm is efficient, it is not differentiable and not learnable. In fact, triangulation algorithms operate on vertices to generate  faces in the form of integer indices. This filtering stage relies on thresholding operations and generates integers that have zero gradients almost everywhere.  In this paper, we propose a novel learnable filtering block that is based on reinforcement learning in order to overcome this disconnection within the gradient chain rule so that the triangulation block can be incorporated in an end-to-end trainable network.

\begin{algorithm}[H] 
\caption{Algorithm of the  triangulation layer. }
\label{alg:trig}
\begin{algorithmic}[1]
\Require{Vertices $v$, state $s$.} 
\Function{Triangulate}{$v,s$}
 \State { tetrahedrons $\gets$ Delaunay$\_$complex(v)}
 \State {$\tau$ $\gets$ PPN($s$)}
 \For{tetrahedron \textbf{in} tetrahedrons}   
 \State{$\tau_{tet}$ = circumsphere$\_$radius(tetrahedron)}
        \If{$\tau_{tet}$ > $\tau$}
        \State {delete(tetrahedron, tetrahedrons)}
        \EndIf
    \EndFor
\State{faces = extract$\_$faces(tetrahedrons)}
\State{Remove duplicated faces from faces.}
    
    \State \Return {faces}
   
\EndFunction
 \Ensure{The set of faces is not empty.}
\end{algorithmic}
\end{algorithm}

Let $t$ be the list of tetrahedrons generated by the first step of the alpha shape algorithm, $f$ their corresponding faces, and $I$ the input image. We treat the filtering stage  as an agent that eliminates from $f$ the faces of the tetrahedrons whose  circumradius is larger than a threshold $\tau$. (See Algorithm~\ref{alg:trig} for more details.)   Let:
\begin{itemize}
    \item The VGN and  MRN be the agent's environment.
    \item The agent's states $s\in \statespace$: the observed image $I$  of the object to reconstruct.
    \item $A$: the action space of the of the agent where  $A(s) = \{\psi (f,\tau_i), i=1, \cdots, n \}$ is the set of possible actions  that the agent can take in each state $s\in \statespace$. Here, $\psi (f,\tau)$ is the action of removing from $f$ the faces that are larger than the threshold $\tau$. In our case, the agent chooses the appropriate threshold $\tau$  out of the $n$ predefined values $\{\tau_1, \dots, \tau_n\}$.
    \item $P(s', s|a)$:  the transition model defined as the probability of transition from state $s$ to state $s'$ given the action $a \in A(s)$. 
    \item  $\pi(s)$: a policy that  maps the agent's state $s \in S$ to actions $a \in A(s)$. 
    
    \item $r$:  the reward function, and 
    
    \item $V_{\pi}(s)$: the  value function of the policy $\pi$, which is  the future reward that the agent would receive by  taking an action in a particular state $s$. 
\end{itemize}
\noi The goal of reinforcement learning is to learn the policy $\pi(s)$ that chooses the optimal threshold $\tau$ among a list of candidate values $\{\tau_i, i=1, \cdots, n\}$ for $s \in \statespace$. This is equivalent to training an agent to choose the best action to perform from a list of candidate actions $A$.

We model the agent using a neural network that approximates the expected reward per action. Let $\theta$ be the parameters of the network. The agent observes an image, which is represented by a state $s$, generates potential rewards per action, and then selects the action with the highest expected reward.
To train the agent, we define the value function $V_{\pi}(s)$ of the policy $\pi$ as the expectation $E$ of the  reward $r$ that the agent gets:
\begin{equation}
\centering
    V_{\pi}(s) = E[r(s,\pi(s),\theta)], s \in \statespace.
    \label{eq:value_function}
\end{equation}

\noi Note here that the reward $r$ is a function of the current state, the policy $\pi$ and the network parameters $\theta$. 
The latter  are constantly updated over time, thus the   value function  $V_{\pi}(s)$ is also constantly updated. We use $V_{\pi}(s)$ as the target ground truth  to train our agent, which we model as a neural network. In contrast to supervised learning where the target values are fixed at the start of the training stage, our target values change as the network parameters $\theta$ gets updated.

We define the reward at training based on the 
fidelity  of the generated mesh to the ground truth mesh.  An optimal policy $\pi^{*}$ is the policy that achieves the maximum mesh reconstruction fidelity for all states $s \in \statespace$.  This mechanism allows the agent to be aware of the reconstruction performance and thus adjust the threshold $\tau$ without relying on the backpropagated gradients.

\vspace{6pt}
\noindent\textbf{Network architecture.} We implement this procedure using a deep neural network composed of \textbf{(1)} a Policy Prediction Network (PPN) and \textbf{(2)} a triangulation layer that joins the VGN and the PPN. The PPN  implements a Q-Network~\cite{mnih2015human}, \ie we approximate an action-value function of Equation~\eqref{eq:value_function} using a neural network. It receives the features extracted by the image encoder, and applies average pooling followed by three fully connected hidden layers with $512$ kernels and a final fully connected layer that outputs the number candidate actions $n$. We use ReLu activation for the hidden layer. Approximating the expected value per action (Equation \eqref{eq:value_function}) makes the proposed agent capable of knowing which filtering action generates accurate and smoother meshes, thus can select the appropriate filtering operation.

\vspace{6pt}
\noindent\textbf{Exploration. } 
The proposed agent disposes of multiple actions that can be applied to the input faces and interact with a  changing environment, \ie the model parameters $\theta$ are continuously updated during the training stage. At the early stages of training, the reward of  an action can change and become higher than the current best reward. The agent should be able to adapt to such change. Thus, we follow an $\explorability$-greedy exploration policy, \ie we learn the greedy policy $a^* = \arg\max_{a} \text{PPN}(s, a,\theta)$ and select a random action with a  probability of $(1-\explorability)$. Instead of optimizing the PPN after each update of $\theta$, we train it periodically, \ie after each $\period$ mini-batch updates of $\theta$, we freeze  $\theta$, compute the mesh quality indicators, and train the PPN. This results in a more stable training. 

At runtime, the proposed triangulation layer receives the vertex positions that have been generated by the VGN, estimates the candidate  faces $f$ using Delaunay complex, and then fetches, for each face, the appropriate filtering actions $\psi(f,\tau)$ using the PPN. These define the filtering operations that need to be applied to the list of candidate faces $f$ in order to only keep  the faces that best represent the geometry of the target  3D object. 
\subsection{Mesh Refinement Network}
\label{sec:mesh_refinement}

The initial 3D mesh inferred by the  Coarse Mesh Generator Block (CGB) may contain holes and artifacts. Although polygon repairing techniques~\cite{ju2004robust,attene2013polygon} can be used to fix these issues, the quality of the reconstruction may not be guaranteed to be at the desired level.

To improve the accuracy of  the reconstructed geometries, we propose to further refine the initially generated mesh so that holes get stitched and artifacts get filtered out. We use a cascade of two refining blocks that use graph convolution-based architectures. The input of the first block is the coarse mesh and the learned features of the input image obtained from the second to last ResNet blocks of the encoder.  We rely on the known camera intrinsic matrix to align the generated vertices with the learned features as in~\cite{wang2018pixel2mesh,gkioxari2019mesh}, \ie we project the generated mesh onto the feature space  using the camera intrinsic matrix.  
The input is then processed with two refinement blocks (Figure~\ref{fig:overview}). Each  block has $\Ncgn$ spectral graph convolution layers with kernels of size $196$. 
Each refinement stage updates  a vertex $v$ of an input mesh $M$ as follows;
$
    \hat{v} = v + \tanh(R(X,M,v))
    \label{eq:vertex_update}
$ where $X$ represents the perceptual features generated by the encoder, and $R(X, M, v)$ is the  offset of the vertex $v$ estimated by the refinement block.

\subsection{Mesh Quality}
\label{sec:mesh_quality}
The process of generating manifold 3D mesh reconstructions is challenging. In fact, even real world meshes often contain self intersections, artifacts, and holes~\cite{attene2013polygon,zhou2016thingi10k}. Learning to generate visually attractive meshes is a long standing ill posed problem as current evaluation metrics fail to evaluate such aspect~\cite{wang2018pixel2mesh,gkioxari2019mesh}. Template based methods such as Pixel2Mesh~\cite{wang2018pixel2mesh} benefit from the clean connectivity of the initial template and are able to generate appealing reconstructions. They, however, are limited to the topology of the template and cannot reconstruct objects of a different topological structure. 

\begin{figure}[t]

    \begin{tabular}{cc}
       \includegraphics[width=.45\linewidth]{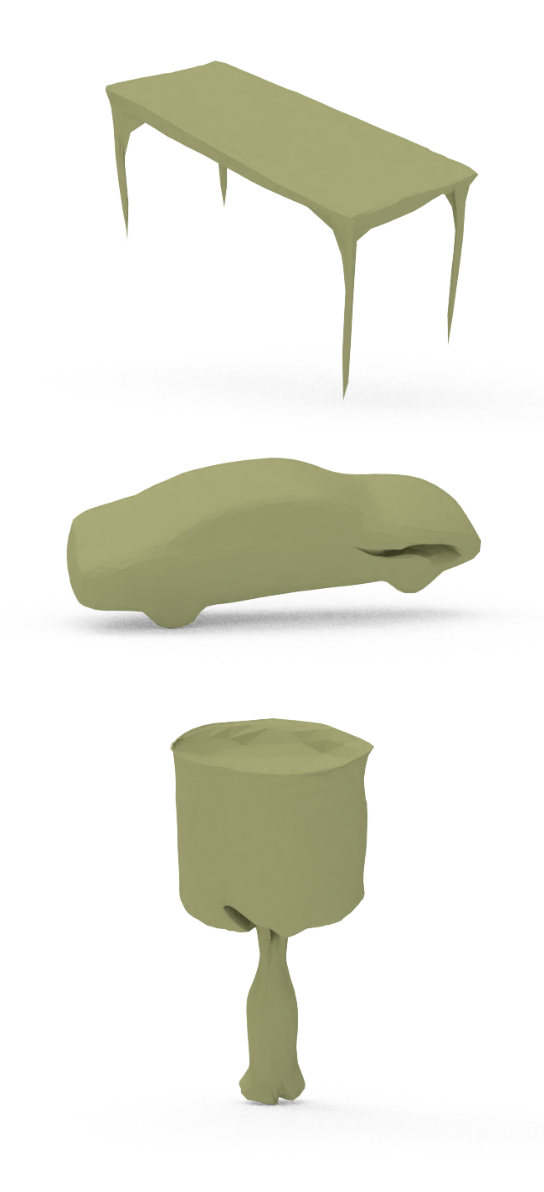}& \includegraphics[width=.45\linewidth]{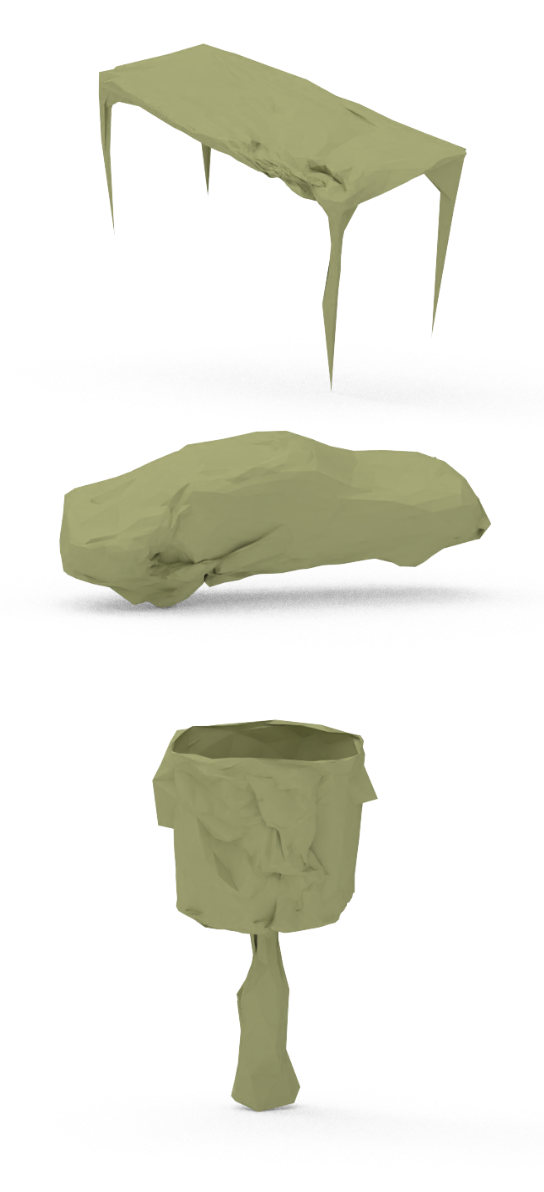} \\
        
        (a) Pixel2Mesh. & (b) Pixel2Mesh (Pretty). \\

    \end{tabular}

\caption{Qualitative comparison between Pixel2Mesh trained (a) with and (b) without  the smooth loss.}

\label{fig:qual_p2m}
\end{figure}

Mesh R-CNN~\cite{gkioxari2019mesh} learns to generate 3D mesh reconstructions by triangulating an intermediate low-resolution volumetric representation. However, the reconstructed meshes exhibit topological errors such as self-intersections,  and may contain non-smooth areas. The latter is mainly due to the use of low resolution voxels but the self intersections are caused during the deformation stage that takes place after the triangulation stage and aims to refine the reconstruction. To prevent such topological noise, Pixel2Mesh~\cite{wang2018pixel2mesh} compares the Laplacian of the reconstructed mesh to that of the initial template and constrains the deformation to preserve the topological aspect portrayed 
by the Laplacian operator. The  version of Pixel2Mesh~\cite{wang2018pixel2mesh} that was trained without the Laplacian constraints (Pixel2Mesh (Pretty)) exhibits non smooth surfaces and face intersections as shown in Figure~\ref{fig:qual_p2m}.

To address these limitations, we propose to create a mesh baseline and use it to constrain the mesh deformation in the refinement stage. This is not intuitive since the baseline has to have the exact topological structure of the target object and has to have a clean connectivity. To do so, we rely on the coarse mesh generated by the Coarse Mesh Generation Block of Section~\ref{sec:coarse_generation}. Since the alpha shape algorithm generates clean connectivity, it is possible to use it as a baseline. However, the vertices generated by the VGN can be noisy and can result in a non-smooth reconstruction. To ensure the smoothness of the baseline, we smooth the resulting mesh of the alpha shape triangulation using Taubin smoothing algorithm~\cite{taubin1995curve}. This results in the shrinkage of leaf parts such as chair legs, which we attempt to overcome by re-triangulating the smoothed vertices and relying on the subsequent Mesh Refinement Network (MRN). Finally, we use the resulting re-triangulated mesh as a baseline to constrain the MRN deformation process.

\subsection{Loss Function }
\label{loss_pretty}
We train the proposed architecture using  a loss function that is a combination of a data term and a regularization term. The data term measures the discrepancy between the reconstructed mesh and the ground truth mesh, while the regularization term  imposes some constraints on the reconstructed meshes to control its quality. Below, we describe in detail these terms.

\vspace{6pt}
\noi\textbf{The data term.} The Chamfer Distance (CMD), one of the commonly used loss functions for comparing two arbitrary point clouds $\mathcal{P}$ and  $\mathcal{Q}$, is defined as: 
\begin{equation}
\centering
    \text{CMD}= \dfrac{1}{|\mathcal{P}|} \sum_{p \in \mathcal{P} }\min \limits_{q \in \mathcal{Q}} 
    \|p-q\|_{2}^{2} + 
    \dfrac{1}{|\mathcal{Q}|}
    \sum_{q \in \mathcal{Q} }^{}\min \limits_{p \in \mathcal{P} } \|p-q\|_{2}^{2}.
    \label{eq:cham_dist}
\end{equation}

\noi One of its limitations is that it depends highly on  the learning rate and on the  initial positions of the generated points. 
Let $\mathcal{N}(p) \in \mathcal{Q} $ be the  nearest neighbours  of a point $p \in \mathcal{P}$, $u \in \{x,y,z \}$ is an arbitrary axis, $u_{p}$ the coordinate of $p$ along the axis $u$, and $j_{0} = \arg\min_{j \in\mathcal{N}(p)} \left\| u_{p} - u_{j}\right\|_{2}$ the closest point from $\mathcal{N}(p)$ to $p$. The gradient  of the CMD along $u_{p}$  is given by:
\begin{equation}
    \frac{\partial \text{CMD}}{\partial u_{p}} = \dfrac{2 u_{p}}{|\mathcal{P}|}  \left\| u_{p} - u_{j_{0}}\right\|_{2} + \sum_{j \in \mathcal{N}_{p}}^{} \dfrac{2 u_{p}}{|\mathcal{Q}|}  \left\| u_{p} - u_{j}\right\|_{2} .
    \label{eq:grad_cmd}
\end{equation}

\noi Those gradients depend on the distance $\left\| u_{p} - u_{j}\right\|$. Their magnitudes are high when $u_{p}$ is far from its corresponding target $u_{j}$, which is the case at the early stages of the training,  and are small at the late stages. 

At the early stages of training, the nearest neighbour selection is not accurate since it depends on the initial positions of the generated points.  At the intermediate stages, the highest value 
$\left\| u_{p} - u_{j}\right\|$ will have the highest 
gradient and the model will be updated to minimize such distance. This is opposite to how the model should be updated. In fact, at the early stages of the training, the model should be updated carefully until the matching between the predicted points and the ground truth points becomes more accurate and less dependent on the initialization. Also, the closest neighbour among the list of neighbours should have the highest gradient. At the later training stages, the gradient vanishes as  $\left\| u_{p} - u_{j}\right\|$ becomes too small. 

To address these issues, Mesh R-CNN~\cite{gkioxari2019mesh} linearly increases the learning rate at the early stages and decreases it at the later stages. The reconstruction task is very sensitive to the choice of the learning rate, which  requires a lot of trial-and-error to ensure convergence.   A bad learning rate update scheduler or a bad learning rate might result in under-fitting or non-convergence. Additionally, the Chamfer distance is scale dependant while objects in the training set may have various scales. Thus, one may require different learning rates per instance. In this paper, we propose a log-based Chamfer distance $\logcmd$ defined as follows: 
\begin{equation}
\centering
\begin{split}
    \logcmd(\mathcal{P}, \mathcal{Q} ) = \sum_{p \in \mathcal{P} }\min \limits_{q \in \mathcal{Q}} 
    \log_{10}(\|p-q\|_{2}^{2} + \mu) + \\
    \sum_{q \in \mathcal{Q} }\min \limits_{p \in \mathcal{P} } \log _{10}(\|p-q\|_{2}^{2} + \mu).
\end{split}
    \label{eq:lof_cham_m_dist}
\end{equation}

\noi  The gradient of $\log_{10} (u+ \mu)$ with respect to $u$, which is equal to $\frac{\log(10)}{u + \mu}$ 

fixes the limitations discussed above. In fact, the gradient of $\logcmd$ is inversely proportional to the distance $u$. Thus, at the early stages of training,  when the matching between points and their nearest neighbours is heavily impacted by the initial positions of the generated points, the gradients have small values. Thus, the model learns and adjusts slowly the positions of the points. However, at the late stages when the matching is more precise, which should be represented by very small distances, the gradients are larger. 
This is equivalent to having a low learning rate at the start of the training and increasing the learning rate as the distance between the generated points and the ground truth points gets smaller. The gradients of the $\logcmd$ adjusts to the scale of the object and  are equivalent of instance-based learning rate scheduling. $\logcmd$ also gives more importance to closer neighbours in contrast to regular CMD, \ie although a point $p$ of a set $\mathcal{P}$ can be the closest neighbour to more than one point of a set $\mathcal{Q}$, the $\logcmd$ gives higher gradient magnitude to the closest neighbour of $\mathcal{Q}$ to $p$. Thus, while updating the point positions, the nearest neighbour of $\mathcal{Q}$ to $p$ is most likely to get closer to $p$. When using regular CMD, the farthest point of $\mathcal{Q}$ to $p$ will have the highest  gradient magnitude while the closest will get the lowest gradient magnitude.

Finally, CMD and $\logcmd$  operate on point-based representations. To adapt them to mesh representations, we convert the estimated meshes into points through uniform sampling using the approach proposed in \cite{osada2002shape}.

\vspace{6pt}
\noi\textbf{The regularization term.} In addition to the data term, we add a regularization term that encourages visually appealing reconstructions. In particular, we observe that by using only the $\logcmd$, the model only focuses on point-to-point distances and omits surface properties. To overcome this, we add a regularization term that encourages smooth surfaces by comparing the Laplacian of the reconstructed mesh to the Laplacian of the mesh of Section~\ref{sec:mesh_quality}. The Laplacian coordinate of a vertex $v_{i}$ is defined as:
\begin{equation}
    LO_{\mathcal{M}}(i) = \sum_{\{j \} \in \mathcal{N}_{v_{i} } }
    w_{i,j}(v_{i} - v_{j}),
    \label{eq:lap_cot1}
\end{equation}

\noi where $ w_{i,j}$ are the cotangent weights, $\mathcal{N}_{v_{i} }$ is the set of first order neighbours of the vertex $v_{i}$ and belonging to mesh $\mathcal{M}$, and $v_{j}$ is a vertex of  $\mathcal{N}_{v_{i} }$ \ie $v_{j}$ shares an edge with $v_{i}$. The regularization is, therefore, defined as the mean squared difference between the Laplacians of the vertices of the two meshes $\mathcal{M}$ and $\mathcal{M_{T}}$:

\begin{equation}
    LR(\mathcal{M},\mathcal{M_{T}}) = \dfrac{1}{|\mathcal{V}|} \sum_{i \in \mathcal{V} } 
    \|LO_{\mathcal{M}}(i) - LO_{\mathcal{M_{T}}}(i) \|_{2}^{2}. 
    \label{eq:lap_cot}
\end{equation}
Here, $\mathcal{M}$ represents a reconstructed mesh and $\mathcal{M_{T}}$ represents the mesh of Section~\ref{sec:mesh_quality}, and $\mathcal{E}$ represents the set of edges of the mesh $\mathcal{M}$.
This regularization term implements the cotangent variant of the Laplacian smoothing~\cite{field1988laplacian} and thus favours the reconstruction of surfaces that are as smooth as the surfaces of the target mesh $\mathcal{M_{T}}$. 

Additionally, we use a normal consistency term to encourage smooth geometry. the normal consistency between two adjacent faces $f_1$ and $f_2$, \ie two faces that share an edge, having $n_{1}$ and $n_{2}$ as normals is defined by:
\begin{equation}
    nc(f_1 , f_2)=  1 - \cos(n_{1}, n_{2}).
\end{equation}

\noi The normal consistency of a mesh $\mathcal{M}$ is, therefore, computed by:
\begin{equation}
    NC(\mathcal{M})=  \sum_{f_1,f_2 \in \mathcal{A}} nc(f_1 , f_2), 
\end{equation}
where $\mathcal{A}$ is the list of adjacent faces of  $\mathcal{M}$.

We also compute the normal consistency between the generated reconstructions and the ground truth, which is defined in terms of angle between the normals $n_{p}$ and $n_{q}$ at points $p$ and $q$, respectively, \ie
\begin{equation}
    \begin{split}
    NL(\mathcal{P}, \mathcal{Q} ) =  \dfrac{1}{|\mathcal{P}|} \sum_{p \in \mathcal{P}, q =  \mathcal{N}(p)   } 
    \cos ({\bf n_{p}},{\bf n_{q}}) +\\ \dfrac{1}{|\mathcal{Q}|} \sum_{q \in \mathcal{Q}, p = \mathcal{N}(q) } 
    \cos ({\bf n_{p}},{\bf n_{q}}).
    \end{split}
    \label{eq:norml_loss}
\end{equation}

\noi In addition to the smoothness terms, we also add a term that penalizes long edges. It is defined as: 
 \begin{equation}
\centering
    EL(\mathcal{M}) = \dfrac{1}{|\mathcal{E}|} \sum_{e \in \mathcal{E} } 
    \|e\|_{2}^{2}. 
    \label{eq:edg_len}
\end{equation}

\noi It is used to prevent the mesh vertices from freely drifting far apart from each other. It also helps the network avoid being stuck in local minima and prevents degenerate reconstructions~\cite{gkioxari2019mesh}.

\vspace{6pt}
\noi \textbf{Overall loss.} We define the overall loss $\text{L}$ as: 
\begin{equation}
    \begin{split}
        \text{L} = \coefcmd  \logcmd(\mathcal{P}, \mathcal{P}_{gt} ) +\coefogcmd  \text{CMD}(\mathcal{P}, \mathcal{P}_{gt} ) \\ +\coeflc LC(\mathcal{M}) +\coefel EL(\mathcal{M}) + \coefnc NC(M\mathcal{M}) +\\ \coefnl NL(\mathcal{P}, \mathcal{P}_{gt} ), 
    \end{split}
 \label{eq:total_loss}
\end{equation}
\noi which is a weighted sum  of the data term and the regularization terms. Here, $\mathcal{P}$ and $\mathcal{P}_{gt}$ are point clouds extracted from $\mathcal{M}$ and $\mathcal{{M}}_{gt}$, respectively.

We train the PPN to approximate the reward of each action. While training, we compute the fidelity to the ground truth meshes using the F1 scores computed  using point-based accuracy,  
\ie a point $p \in \mathcal{P}$ is considered accurate if at least one point $q \in \mathcal{P}_{gt}$ is found within a sphere centered at $p$ and of radius $10^{-4}$. 
Thus, the reward used to train the PPN is defined as: 
\begin{equation}
    r(\mathcal{M}, \mathcal{{M}}_{gt} ) = F_{1} (\mathcal{P}, \mathcal{P}_{gt} ).
    \label{eq:reward}
\end{equation}

\section{Experiments}
\label{sec:results}
We evaluate the performance of the proposed framework following several 
evaluation protocols, which we detail in (Section~\ref{sec:evaluation_protocols}). Section~\ref{subsec:training} discusses the training details. We then evaluate and compare the  performance of the proposed model to state-of-the-art methods using (1) ShapeNet~\cite{chang2015shapenet} (Section~\ref{sec:evaluation_shapenet}) and   Pix3D~\cite{sun2018pix3d} (Section~\ref{sec:pix3d}), which are public datasets, and (2) real-world images from the Internet (Section~\ref{sec:inthewild}). We also perform an ablation study to evaluate the contribution of the individual components of the proposed approach (Section~\ref{sec:ablation_study}).

\subsection{Evaluation metric and protocols}
\label{sec:evaluation_protocols}
Several measures have been used in the literature~\cite{gkioxari2019mesh,wang2018pixel2mesh,tang2019skeleton,pan2019deep} to assess the performance of 3D object reconstruction algorithms from monocular images.   The point-based Chamfer Distance (CD) is the most commonly used assessment measure to compare two arbitrary shapes, because of its rotation-invariant and topology-independent properties. It generates point clouds from the 3D mesh before assessing the nearest neighbor $L_{2}$ distance between the two sets. However, as stated in~\cite{wang2018pixel2mesh,gkioxari2019mesh}, there is no strong correlation between minimizing a surface-based Chamfer distance and the high reconstruction fidelity, especially when it comes to the visual aspect. Gkioxari \etal~\cite{gkioxari2019mesh} introduced two models, the Best and the Pretty, and suggested that the model with the best (\ie lowest) Chamfer distance performs poorer in  terms of visual quality. In this work, in addition to the proposed model, hereinafter referred to as \emph{Smooth}, we also introduce two other models: the Best and the Pretty.
\begin{itemize}
    \item The  Smooth model (Section \ref{sec:smooth}) is trained following a newly introduced training protocol to ensure reconstructions of high visual quality.
    
    \item  The Best model (Section~\ref{sec:best})  is trained solely on Chamfer distance following Mesh R-CNN~\cite{gkioxari2019mesh}'s Best training protocol.
    
    \item The Pretty model (Section~\ref{sec:pretty}) is trained following Mesh R-CNN~\cite{gkioxari2019mesh}'s Pretty training protocol.
\end{itemize}

\noi Also, the Chamfer distance is not a metric and is scale dependent. Thus, to ensure fairness in the comparison, we evaluate the proposed method using four protocols that are  used in the literature:
\begin{itemize}
    \item\textbf{Pixel2Mesh protocol~\cite{wang2018pixel2mesh}}  states that meshes should be rescaled by a factor of $0.57$ to align them with the rendered images. Also, it uses the Chamfer distance between uniformly sampled points on the reconstructed and ground-truth 3D models and the F1 score as a measure of the reconstruction accuracy. With this measure, a prediction is considered accurate if a ground truth point cloud is found within a sphere of a certain radius $r$. The induced measure is denoted by $F_1^{r}$, with $r\in\{0.1, 0.2\}$.

    \item \textbf{Mesh R-CNN protocol~\cite{gkioxari2019mesh}.} Mesh R-CNN~\cite{gkioxari2019mesh} proposed a new test set that  considers only the 3D CAD models of  ShapeNet that are of  high genus. The training and testing data are augmented with $19$ more view angles compared to only $5$ in the originally generated $24$ views used by Pixel2Mesh. The meshes are scaled so that the longest edge of their bounding box is of  length $10$. The accuracy is then measured using the L1 distance between the ground truth and the reconstructed models. Finally, the  cosine similarity is used to assess the smoothness of the generated surfaces. 

    \item \textbf{TMNET protocol~\cite{pan2019deep}.} TMNET~\cite{pan2019deep} evaluates the accuracy of the reconstruction by first  aligning the reconstructed 3D model to the canonical view of the groundtruth model in ShapeNet, using the ground truth rotation matrix. The reconstruction error is then measured using the Iterative Closest Point (ICP).

    \item \textbf{Skeleton Bridge protocol~\cite{tang2019skeleton}.} This protocol reports the Chamfer Distance on $10$ classes of ShapeNet~\cite{chang2015shapenet}. Unlike other protocols that average the reconstruction error across all classes, this protocol reports the reconstruction error on each individual class. 
   
\end{itemize}

\subsection{Training details}
\label{subsec:training}
We describe in this section the training details of the three versions of the proposed model.

\subsubsection{The full (Smooth) model}
\label{sec:smooth}
We train the proposed full (Smooth) model on ShapeNet and Pix3D data sets.

\vspace{6pt}
\noi\textbf{Training on ShapeNet.} First we train the VGN of the proposed Smooth model using Adam optimizer with a learning rate of $3 \times 10^{-5}$ for $100$ epochs. We use $\logcmd$ as loss function between the predicted vertices and ground truth point cloud. We then freeze the VGN and train the PPN for one epoch using RMSProp~\cite{hinton2012neural}  as optimizer with a learning rate of $10^{-4}$. We fix the explorabilty rate at $90\%$. Finally, we periodically train the PPN and MRN as explained in Section~\ref{sec:policy_prediction} and by adopting RMSProp~\cite{hinton2012neural} and Adam, respectfully. We use a learning rate of $10^{-4}$ for both optimizers. We set $\explorability = 10\%$ and $\period = 2$, $\coefogcmd = 1, \coefel = 0.15$, $\coeflc = 0.5$, $\coefnc = 10^{-3}$, $\coefnl = 10^{-4}$, and $\nu = 10^{-4}$. We found that  $ \tau \in \{ 0.05, 0.085, 0.11 \}$ performs best in terms of visual aspect and reconstruction fidelity. %

\subsubsection{The proposed Best model}
\label{sec:best}
The proposed Best model uses a different architecture than the Smooth model. It uses  InceptionNet network architecture~\cite{szegedy2016rethinking} to map an input  RGB image of size $224 \times 224$ to a latent representation (feature vector) of size $1536$. The learned features are then processed by three 3D point-regression blocks. Each of the  regression blocks is composed of three fully connected layers, each of which is preceded by a batch normalization and followed by a nonlinear activation function. The first two layers use a Rectified Linear unit (ReLu), while the last layer uses the hyperbolic tangent function (tanh). 

The \textbf{first} regression block estimates, from the latent representation,  the  coarse 3D shape represented as a set of $750$ points. The \textbf{second} block infers a more detailed point cloud representation. The output of the first block is merged, by concatenation,  with the learned features and fed to the second block, which produces a detailed 3D geometry of size  $1500$ points. Similarly, we merge the feature vector with the output of the second block and feed them to the last block to generate a final output of $3$K vertices and $1$K faces. The The entire pipeline in trained in an end-to-end fashion using a batch size of $128$ and Adam optimizer with a learning rate of $10^{-4}$. We use CMD as a loss function.

\subsubsection{The proposed Pretty model}
\label{sec:pretty}
The proposed Pretty model has a slightly different architecture than our proposed Smooth model. The VGN of the Pretty model outputs $256$ points, instead of $1536$ for the Smooth models, and the MRN contains subdivision operations that perform edge-based graph unpooling thus subdivides the faces of the mesh by a factor of five~\cite{wang2018pixel2mesh}. The PPN of the Pretty model implements a shallower network composed of two fully connected layers, compared to four in the PPN of the smooth model.

To speed up the training of the Pretty model, we initiate the training process  with a  warm-up phase of $20$ epochs. During this phase,  we solely train the point cloud section of the geometry generation network using only the data term of the loss function  (Equation ~\ref{eq:lof_cham_m_dist}). We use Adam as optimizer with a learning rate of $3\times 10^{-5}$, a weight decay of $10^{-6}$, and a decay factor of $0.95$. After the warm-up phase, we train the entire network in an end-to-end fashion using Adam with a learning rate of $2 \times 10^{-6}$, $10^{-7}$, and  $10^{-5}$ for the MRN, VGN, and PPN respectively, and an explorability rate of $20\%$, which we  decay  by a factor of $0.95$ after every $200$ optimisation steps while making  sure that it does not drop below $1\%$. In practice, we found that using $24$  filtering actions with $\tau \in \{0.15+i/50, i \in \{-12, \cdots, 11\}\} $ provides the best results. We set $\coefogcmd = 1, \coefel = 0.2$, $\coeflc = 0$, $\coefnc = 0$, and $\coefnl= 0$ to align with Pretty training protocol of Mesh R-CNN~\cite{gkioxari2019mesh}. The training ends when the performance on the validation set starts to drop, which happened after $73$ epochs. We set $\nu = 10^{-4}$.

\subsection{Performance on ShapeNet}
\label{sec:evaluation_shapenet}
We evaluate the proposed method on the ShapeNet benchmark~\cite{chang2015shapenet}, which contains textured 3D models from $55$ shape categories. Following the protocols described in \cite{fan2017point}, we use a  subset composed of $35,011$ models from $13$ shape categories. We render from each 3D model  $24$ views of a white background and  of size $137\times137$. This results in $840,264$ images, each one is paired with its corresponding 3D model. 

{
\begin{table*}[t]
\caption{Reconstruction error on ShapeNet following Mesh R-CNN protocol, which is scale invariant. We compare our method to state-of-the-art methods under various training protocols.}
\resizebox{\textwidth}{!}{
\begin{tabular}{@{}lrrrrr@{}c@{}rrrrr@{}ccccc@{}}
\toprule& \multicolumn{5}{c}{Full Test Set} & \phantom{abc}& \multicolumn{5}{c}{Holes Test Set}&& & \textbf{$|V|$} & \textbf{$|F|$} &Time(ms) \\\cmidrule{2-6}  \cmidrule{8-12}& CD $\downarrow$ & Normal $\uparrow$ & F$_{1}^{0.1}$ $\uparrow$ & F$_{1}^{0.3}$ $\uparrow$ & F$_{1}^{0.5}$ $\uparrow$ && CD $\downarrow$ & Normal $\uparrow$ & F$_{1}^{0.1}$ $\uparrow$ & F$_{1}^{0.3}$ $\uparrow$ & F$_{1}^{0.5}$ $\uparrow$ \\
\midrule
Pixel2Mesh		& 0.265 &0.729 & 29.9 & 76.2 &89.0&&0.273 & 0.733 & 30.8& 76.5 & 88.9 && &2466 $\pm$0 &4928 $\pm$0  &50$\pm$0 \\
\midrule
Mesh R-CNN (Best) 	&0.133 & 0.729 & 38.8 & 86.8 & 95.1 &&\textbf{0.130} & 0.725 & 41.7 &
\textbf{86.7} & \textbf{94.9} && &2291 $\pm$903 &4595 $\pm$1814 & -  \\
Pixel2Mesh (Best) 	&0.132 & 0.711 & 38.3 & 86.5 & 95.1 &&0.138 & 0.705 & 40.0 &
85.5 & 94.4 && &2562 $\pm$0 &5120 $\pm$0 & -  \\
 Ours (Best)  &\textbf{0.118} & \textbf{0.731} &  \textbf{51.4} & \textbf{88.4}  &   \textbf{95.5}&& 0.133  & \textbf{0.767} & \textbf{51.3}  & 86.5  & 94.5 && &3000 $\pm$0 &1000 $\pm$0 &69.2$\pm$0.68 \\
\midrule
Mesh R-CNN (Pretty) &0.171 & 0.713 & 35.1 & 82.6 & 93.2 && 0.171 & 0.700 & 37.1 & 82.4 & 92.7  && &1896 $\pm$928 &3795 $\pm$1861 &679$\pm$31.8\\
Pixel2Mesh (Pretty) 	&0.175 & 0.727 & 34.9 & 82.3 & 92.9 &&0.196 & 0.685 & 34.4 &
79.9 & 91.4 && &2562 $\pm$0 &5120 $\pm$0 & -  \\
Ours (Pretty)  &\textbf{0.155} & \textbf{0.729}&  \textbf{36.6}& \textbf{84.0} &  \textbf{93.8} && \textbf{0.152}  & \textbf{0.708} &  \textbf{38.0}&  \textbf{83.6} & \textbf{93.4}  && &3490 $\pm$684 &6977 $\pm$1198  &183$\pm$103\\
\midrule
Ours (Smooth) &0.187 & 0.734 &33.3 & 80.3 & 92.1 && 0.192 & 0.721 & 35.3 & 80.5 & 91.7  && &908$\pm$283  &1854$\pm$611  &89.3 $\pm$ 0.714 \\

\bottomrule
\end{tabular}
}

\label{tab:prot2_shapenet}
\end{table*}
}
\begin{table}
\centering
\caption{ \label{tab:prot1} Evaluation on ShapeNet dataset following Pixel2Mesh protocol. We report the CD and $F1$ scores.}

 {
\begin{tabular}{@{}l  c  c  c@{}}
    \toprule
   \multirow{2}{*}{\textbf{Model}}&             \multicolumn{1}{c}{\multirow{2}{*}{\textbf{CD $\downarrow$}}}&                   \multicolumn{2}{c}{\textbf{F1 $\uparrow$}} \\
    \cline{3-4} 
          & & $\tau$& $2\tau$\\
        \midrule
        N3MR~\cite{kato2018neural}  &           2.629&           33.80&       47.72\\
        
        D-R2N2~\cite{choy20163d}   &      1.445&       39.01&         54.62\\
        
        PSG~\cite{fan2017point}   &      0.593&       48.58&         69.78\\

        Pixel2Mesh~\cite{wang2018pixel2mesh}   &       0.591&           59.71&       74.19\\
        
        MVD~\cite{smith2018multi}   &       -&           66.39&       -\\
        GEOMetrics~\cite{smith2019geometrics}   &       -&           67.37&       -\\
        \midrule
        Mesh R-CNN (pretty)~\cite{gkioxari2019mesh}    &     0.391&       69.83&    81.76\\

        Ours (Pretty)           &      \textbf{0.364}&        \textbf{71.60}       & \textbf{83.06}\\
              \midrule
        Ours (Smooth)           &      0.402&      67.95      & 80.28\\

   \bottomrule
 \end{tabular}
}

\end{table}

\vspace{6pt}
\noindent\textbf{Quantitative evaluation.} Tables~\ref{tab:prot2_shapenet} and~\ref{tab:prot1} 
compare the performance of the proposed model to volumetric methods such as  D-R2N2~\cite{choy20163d} and MVD~\cite{smith2018multi},  point-based methods such as PSG~\cite{fan2017point}, mesh-based methods such as  N3MR~\cite{kato2018neural},  Pixel2Mesh~\cite{wang2018pixel2mesh} and GEOMetrics~\cite{smith2019geometrics}, which are template based, and  Mesh R-CNN~\cite{gkioxari2019mesh}, which  reconstructs meshes of arbitrary topology but uses voxels as an intermediate  representation.

TMNet~\cite{pan2019deep} adjusts the topology of an initial template, a sphere in this case, by using face pruning operations. In theory, it is capable of generating objects of complex topological structures.  Table~\ref{tab:TMNet} compares the performance of our method to TMNet~\cite{pan2019deep}. As described in TMNet~\cite{pan2019deep} paper,  we compare the Chamfer distance computed between unscaled ShapeNet~\cite{chang2015shapenet} 3D models in their canonical view and 
reconstructions after applying Iterative Closest Point (ICP). For TMNet~\cite{pan2019deep} 
and Pixel2Mesh~\cite{wang2018pixel2mesh}, we report the performance reported by Pan \etal~\cite{pan2019deep}. We also re-evaluate Pixel2Mesh~\cite{wang2018pixel2mesh} and report in Table~\ref{tab:TMNet} its performance with and without applying ICP. The performance of Pixel2Mesh is very different from the one reported in  TMNet~\cite{pan2019deep}. 

\begin{table}[t]
\center{
\caption{We report the Chamfer distance (in millimeters) on ShapeNet following TMNet~\cite{pan2019deep} evaluation protocol. * refers to the performance reported by TMNet~\cite{pan2019deep}. The lower is the error, the better. "ICP" stands for Iterative Closest Point.}
\begin{tabular}{@{}cccccc @{}}
    \toprule
        \textbf{P2M}& \textbf{TMNet} &  \textbf{P2M} & \textbf{P2M}& \textbf{Ours}& \textbf{Ours}\\
        \textbf{(ICP)*}& \textbf{(ICP)*} &  & \textbf{(ICP)}& \textbf{(Pretty)} & \textbf{(ICP)}\\
    \hline 
        $3.531$ & $2.277$  &$1.884$ & $1.195$ & $1.076$ &\textbf{$\textbf{0.906}$}\\
    \bottomrule
    \label{tab:TMNet}
\end{tabular}
}
\end{table}

Similar to Mesh R-CNN~\cite{gkioxari2019mesh}, Skeleton Bridged~\cite{tang2019skeleton} relies on an intermediate volumetric representation of size $128^{3}$, which  is much higher than the $48^{3}$ used in mesh R-CNN~\cite{gkioxari2019mesh}. It also uses a different train/test split. Also, although not explicitly stated in their paper, according to their publicly available code and weights they train a model per category and only consider $10$ categories of ShapeNet benchmark~\cite{chang2015shapenet}. This is different from  previous methods that train one model on all the $13$ categories.

Table~\ref{tab:skeleton} compares the Chamfer distance of our Smooth mode to Skeleton Bridged~\cite{tang2019skeleton} following their evaluation protocol. Our model (Smooth) has been trained on multiple categories and outperforms the per-category trained models of Skeleton Bridged~\cite{tang2019skeleton} on eight of the ten categories we considered.

\begin{table}
\caption{\small{Quantitative comparison with Skeleton Bridged~\cite{tang2019skeleton} on the ShapeNet benchmark~\cite{chang2015shapenet}.  We report the Chamfer distance, in millimeters, of our model (Smooth) trained on the $13$ categories of ShapeNet~\cite{chang2015shapenet}. The lower the error the better is the result.}}
\centering
\begin{tabular}{l|c|c}
\toprule
Category  &Skeleton&Ours (Smooth)\\
\hline
Plane &$1.364$&\textbf{$\textbf{1.044}$}\\
Bench &$1.639$&\textbf{$\textbf{0.910}$}\\
Chair &$\textbf{1.002}$&$1.342$\\
Firearm &$1.793$&\textbf{$\textbf{1.142}$}\\
Table &$1.321$&\textbf{$\textbf{1.153}$}\\
Car &\textbf{$\textbf{0.717}$}& $0.925$\\
Cabinet &$1.708$&\textbf{$\textbf{1.072}$}\\
Couch &$1.350$&\textbf{$\textbf{1.232}$}\\
Lamp &$3.639$&\textbf{$\textbf{2.673}$}\\
Watercraft&$1.597$&\textbf{$\textbf{1.490}$}\\
\hline
Mean &$1.512$&\textbf{$\textbf{1.298}$}\\
\hline
Monitor&-&$1.339$\\
Speaker & - & $1.776$ \\
Cellphone & -& $0.773$ \\

\bottomrule
\end{tabular}

\label{tab:skeleton}
\end{table}

The results in Table~\ref{tab:prot2_shapenet} have been obtained using the evaluation protocol of Mesh R-CNN~\cite{gkioxari2019mesh}.  As noted in~\cite{gkioxari2019mesh,wang2018pixel2mesh}, the quantitative evaluation may be misleading as it does not take into account the connectivity between the generated vertices. Gkioaxari \etal~\cite{gkioxari2019mesh} propose two models. One (Best) optimized solely using Chamfer distance and another one (Pretty) adds a regularization term. They claim that their Best model generates degenerate meshes of intersecting faces and inconsistent normals. Similarly, we report in Table~\ref{tab:prot2_shapenet} the performance of three models. Our (Best) model,  which has not been regularized, thus the mesh faces are allowed  to move freely,  has a very high accuracy based on the adopted evaluation metrics and outperforms the state-of-the-art  on most of the  evaluation metrics. However, it generates degenerate reconstructions. In contrast, the proposed Pretty model performs worse in terms of the Chamfer distance as it imposes more structure,  \eg edge length, on the reconstructed mesh. Finally, our Smooth model imposes higher regularization on the reconstructed meshes, \eg Laplacian smoothness and normal consistency, to strike a balance between reconstruction accuracy and manifoldness. From this experiment, we observe that the more structure we impose on the final mesh, the better will be the visual aspect of the reconstructed mesh while the Chamfer distance worsens. This suggests  that the Chamfer distance is not a suitable metric for quantifying the visual quality of 3D  reconstruction.

\vspace{6pt}
\noi\textbf{Computation time.} Table \ref{tab:prot2_shapenet} reports the average computational time required to reconstruct one 3D model when ran on a i9 processor with a Nvidia Titan RTX GPU card. The average is computed on 1000 randomly selected models from the Shapenet~\cite{chang2015shapenet} test set. For Pixe2Mesh, we indicate the computational time reported by the authors since it was built on Tensorflow and a fair comparison with our meothod and Mesh R-CNN~\cite{gkioxari2019mesh} that use Pytorch is not possible. As one can see, our method (Pretty), although it generates higher mesh resolutions and better accuracy (see Table \ref{tab:prot1}),  is significantly faster than Mesh R-CNN, which uses volumetric grids. Note that we mainly compare our method to Mesh R-CNN as it is open source and recovers 3D models with varying topological structures unlike template based methods such as Pixel2Mesh, which are limited to the reconstruction of 3D models that have the same topology as the template. Our Smooth model  generates lower resolution meshes at a  significantly faster reconstructions, therefore, fair comparison on computational time with state-of-the-art method is not possible.

\begin{figure*}[!ht]
\resizebox*{!}{0.9\textheight}{
    \begin{tabular}{ccccc}
        \includegraphics[width=.19\textwidth]{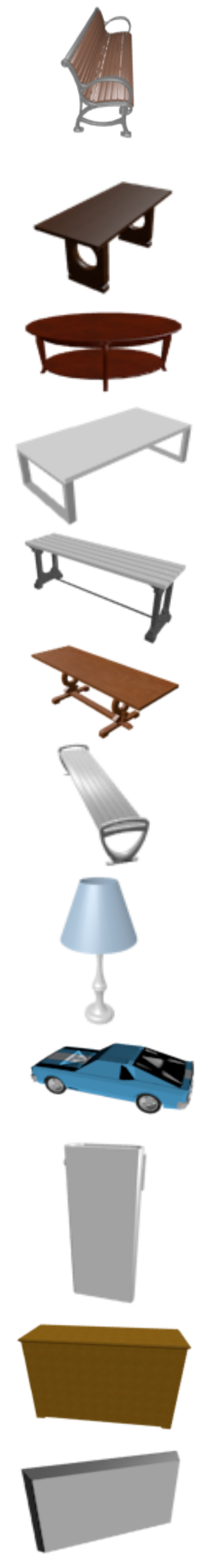} & \includegraphics[width=.19\textwidth]{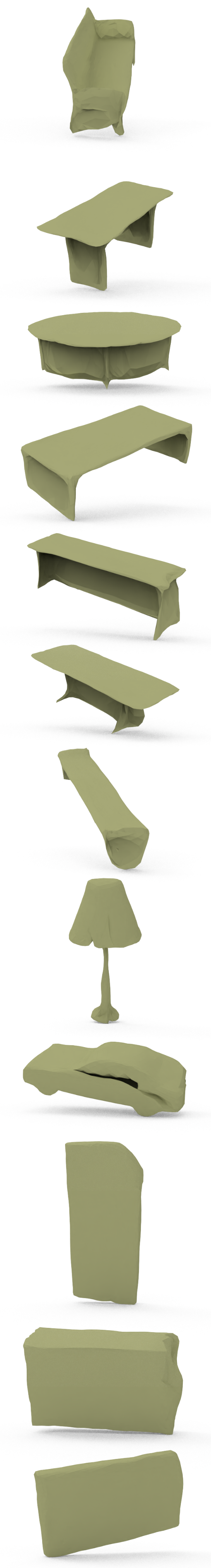}& \includegraphics[width=.19\textwidth]{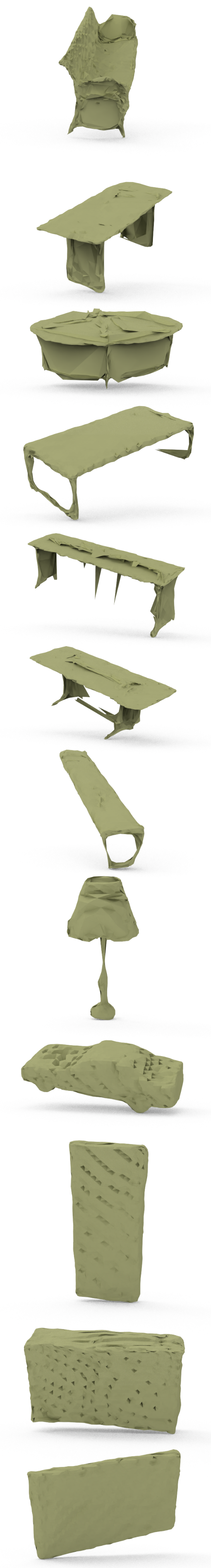}& \includegraphics[width=.19\textwidth]{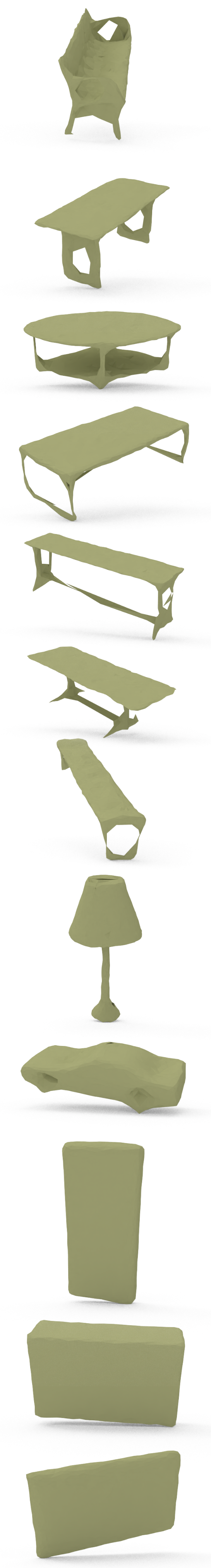}& \includegraphics[width=.19\textwidth]{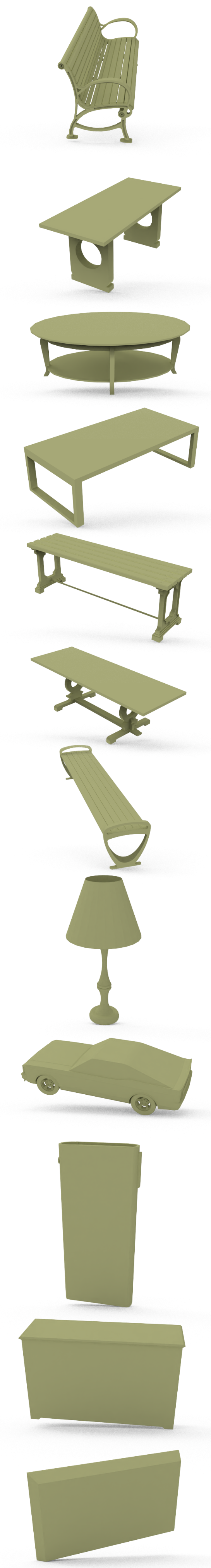}\\
        
       (a) Input. & (b) Pixel2Mesh. & (c) Mesh R-CNN. & (d) Ours (Smooth). & (e) Ground truth.
    \end{tabular}
}

\caption{Comparison on ShapeNet of our approach (Smooth) to Pixel2Mesh and Mesh R-CNN.}
\label{fig:ResultsShapeNettab}
\end{figure*}

\vspace{6pt}
\noindent\textbf{Qualitative evaluation.}
Figure~\ref{fig:ResultsShapeNettab} compares the visual quality of the 3D shapes reconstructed with (1) our method (Smooth), (2) Pixel2Mesh,  and (3) Mesh R-CNN. Our proposed approach reconstructs 3D models that are of higher fidelity to the input image
than state-of-the-art methods. Pixel2Mesh generates smooth reconstructions but,  since it uses template deformation, it cannot reconstruct objects with a topology that is different from the template (see the first seven rows of Figure~\ref{fig:ResultsShapeNettab}). While Mesh R-CNN can reconstruct arbitrary topologies, the use of coarse voxel grids limits its ability to  accurately reconstruct fine structures such as the chair legs~\cite{gkioxari2019mesh}.

\begin{figure}[ht]
\centering
        \begin{tabular}{@{}c@{}c@{}c@{}c@{}c@{}}
            \includegraphics[width=0.2\linewidth]{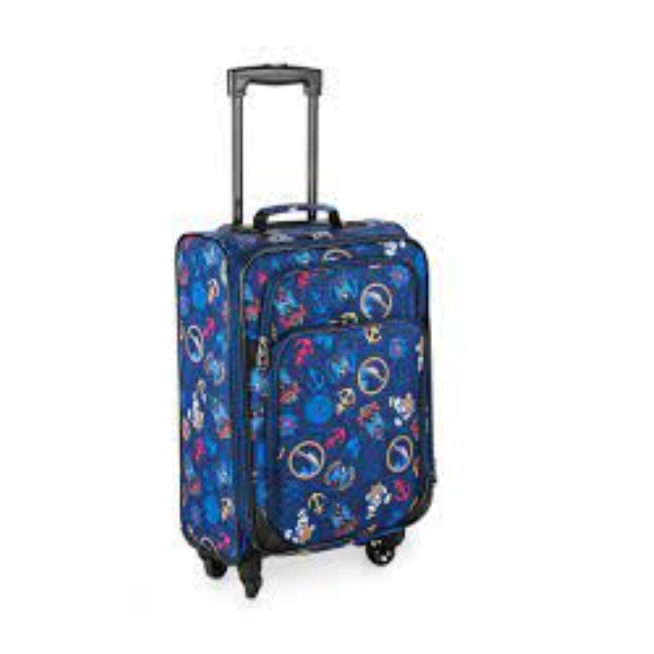} &
            \includegraphics[width=0.2\linewidth]{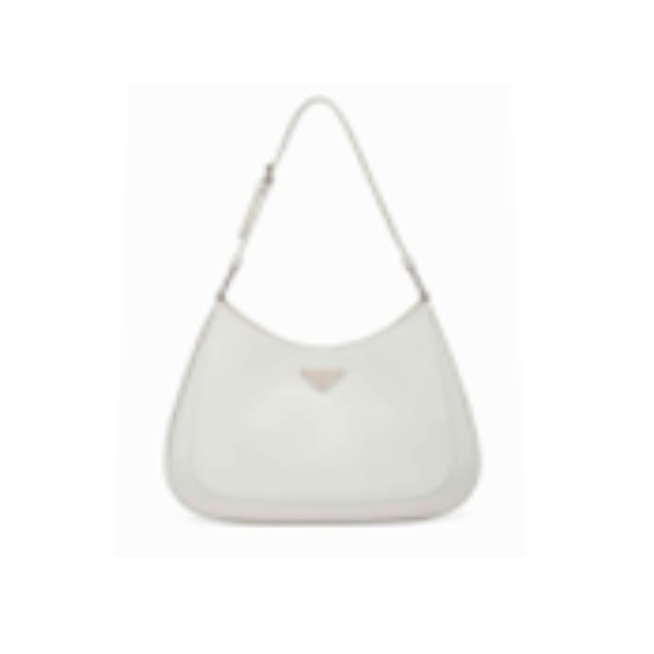} & 
            \includegraphics[width=0.2\linewidth]{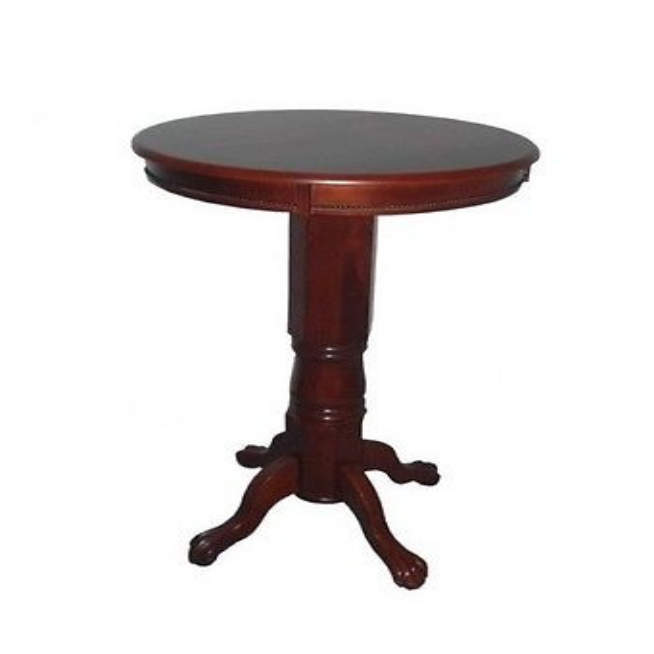}  & 
            \includegraphics[width=0.2\linewidth]{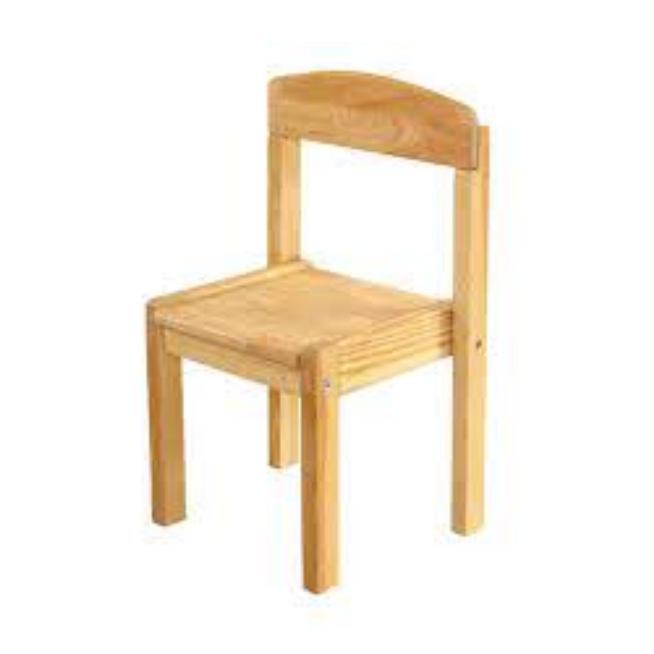}  & 
            \includegraphics[width=0.2\linewidth]{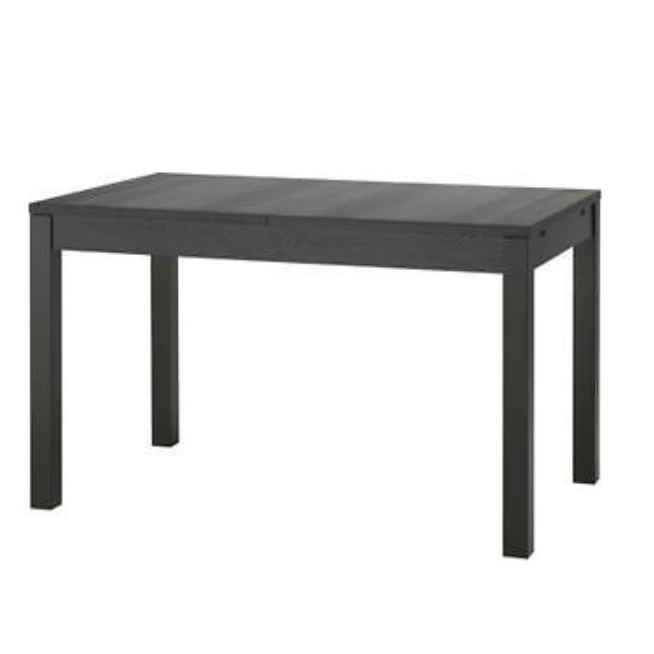}   \\
            \multicolumn{5}{c}{{(a) Input images.}}\\
            \includegraphics[width=0.2\linewidth]{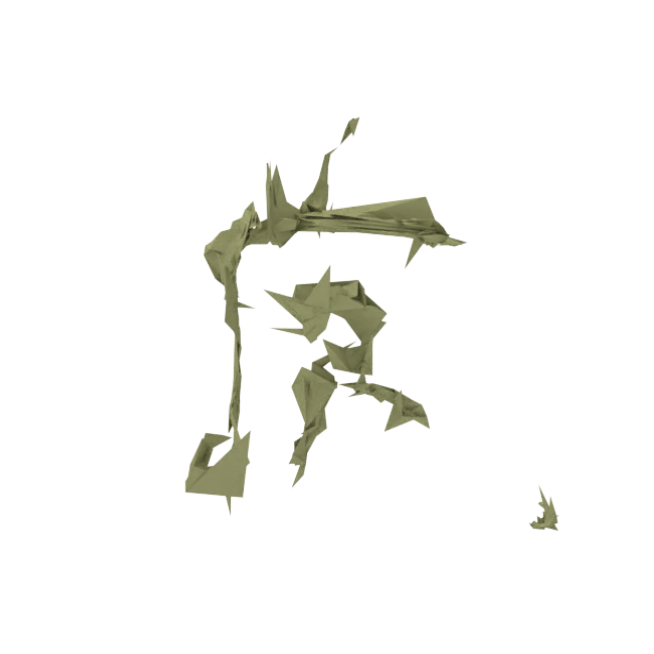} & 
            \includegraphics[width=0.2\linewidth]{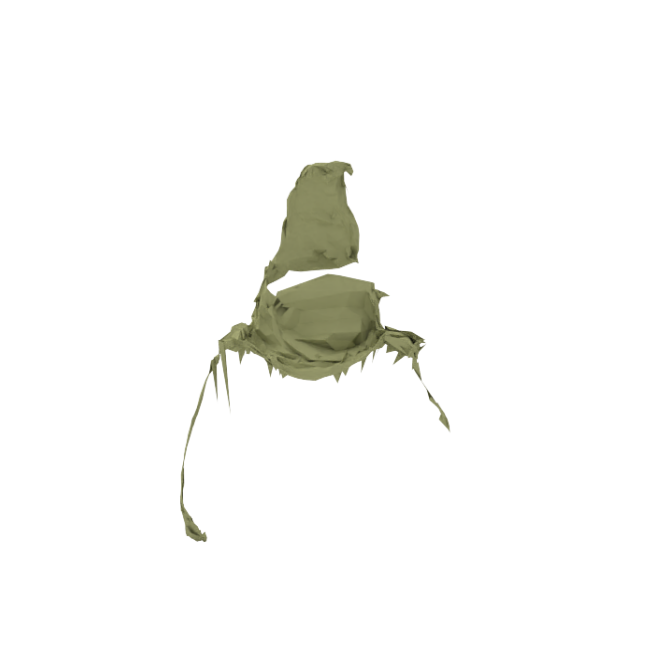} & 
            \includegraphics[width=0.2\linewidth]{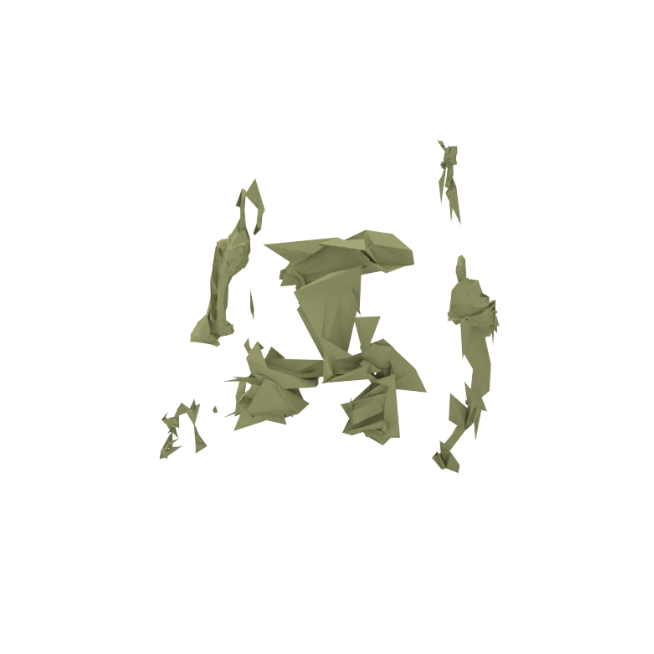} & 
            \includegraphics[width=0.2\linewidth]{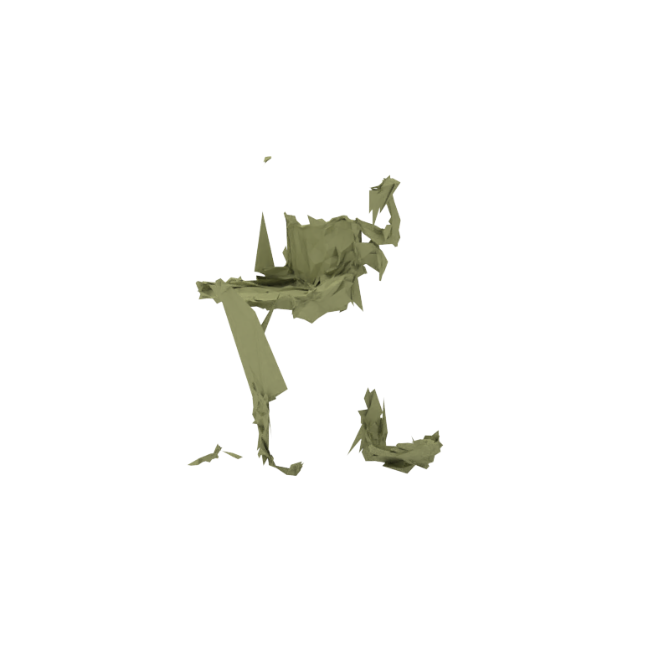} & 
            \includegraphics[width=0.2\linewidth]{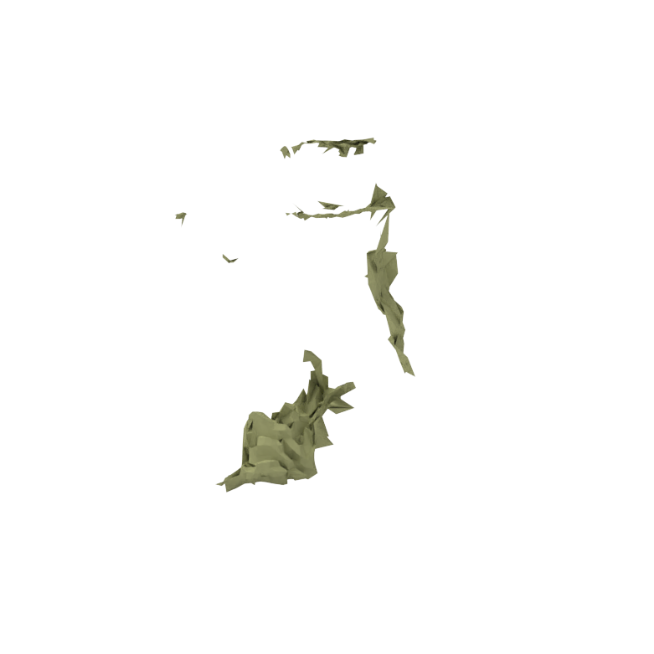} \\
           \multicolumn{5}{c}{{(b) 3D models reconstructed using Mesh R-CNN.}}\\
           \includegraphics[width=0.2\linewidth]{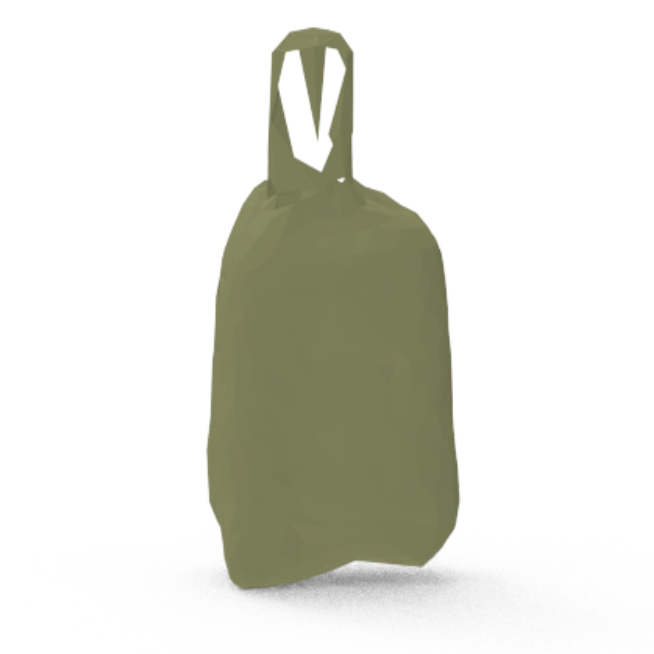} & 
           \includegraphics[width=0.2\linewidth]{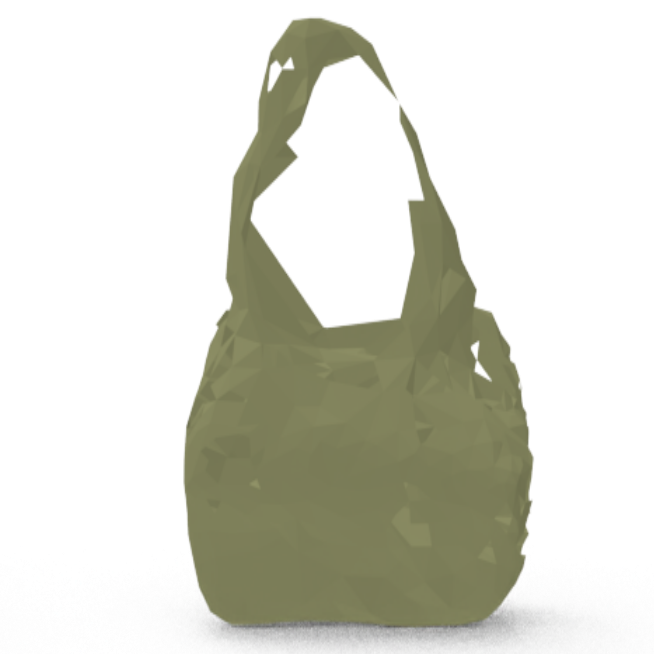} & 
           \includegraphics[width=0.2\linewidth]{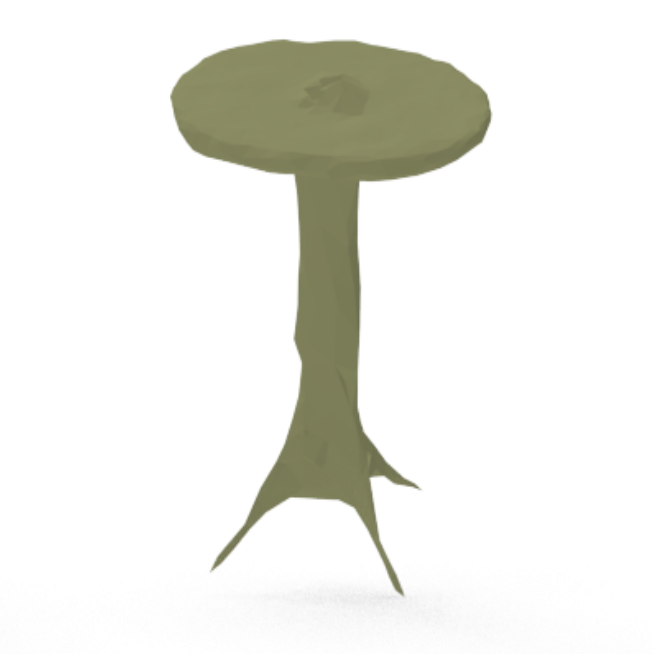} & 
           \includegraphics[width=0.2\linewidth]{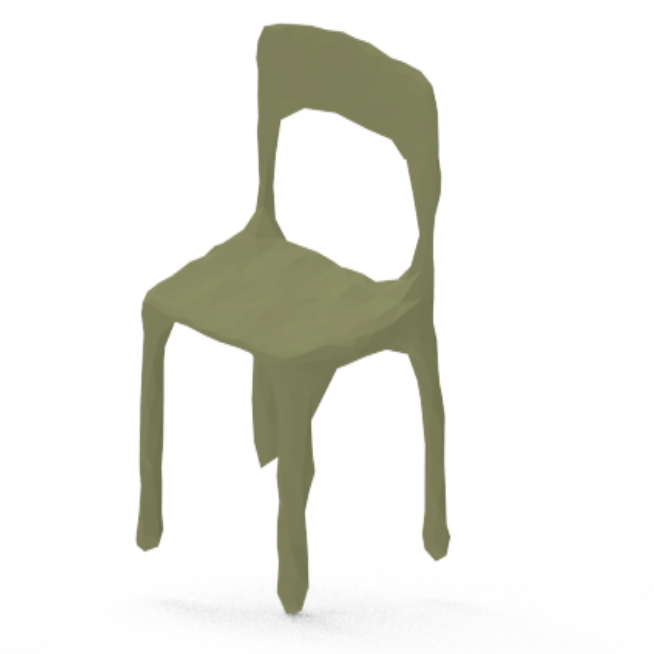} & 
           \includegraphics[width=0.2\linewidth]{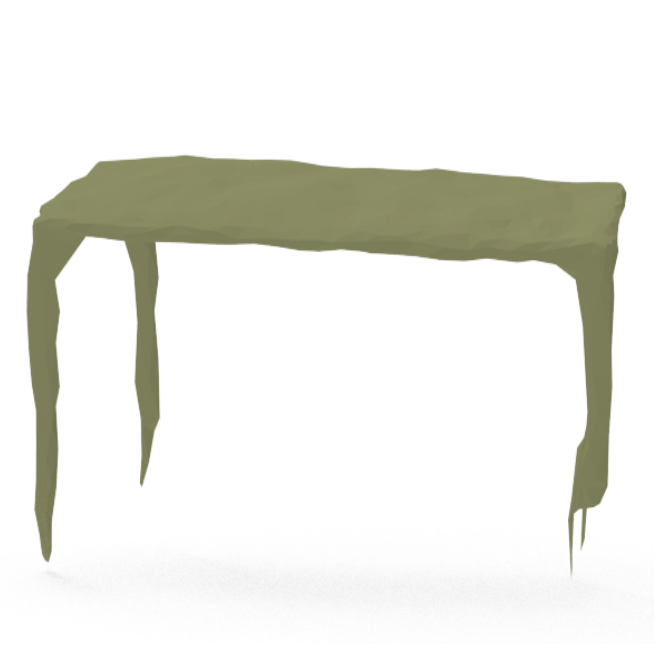} \\
           \multicolumn{5}{c}{{3D models reconstructed with our method (Smooth)}.}
        \end{tabular}
\caption{{Qualitative comparison on in-the-wild images. We compare the visual quality of our method to Mesh R-CNN when tested on real-world images from the Internet, including out-of-category images. Both models have been trained on ShapeNet~\cite{chang2015shapenet}, which is composed of synthetic data. }
}
\label{fig:wild}
\end{figure}

By generating vertices in the appropriate positions and relying on a learnable triangulation layer, our method reconstructs objects with  high fidelity compared to the state-of-the-art. This demonstrates that the proposed method can successfully recover 3D models of complex topology and is computationally more efficient than volumetric techniques.

\subsection{Performance on Pix3D}
\label{sec:pix3d}

Pix3D consists of $10,069$ real images of $395$  CAD models.  Compared to ShapeNet, it is more challenging due to  the misalignment between the images and their corresponding ground truth 3D models. The images also include  occlusions and complex lighting settings.  Since Pix3D contains a small number of images, we first train the proposed Smooth model on ShapeNet~\cite{chang2015shapenet} and then fine-tune it on Pix3D dataset.  We also remove the MRN, since it uses the camera intrinsic matrix. We set $\coeflc = 0.7$ to encourage smoother meshes. We initiate the training with a learning rate of $3 \times 10^{-5}$ and decay it every epoch by a factor of $0.995$ for the first $100$ epochs and $0.999$ for the remaining $200$ epochs. We set the explorability rate to $0.5$ and decay it by a factor of $0.99$ after each optimization step. The rest of the parameters are set the same way as the training on ShapeNet. We adopt the same training-testing split as Mesh R-CNN.

Unlike ShapeNet models, Pix3D~\cite{sun2018pix3d} CAD models are not normalized. This  causes depth/scale ambiguity. To overcome this ambiguity, we normalize the CAD models to fit within a bounding box of size $[-1.45,1.45]^{3}$ to align them with ShapeNet models. At runtime, our model generates normalized models in the world coordinates. We note that the scale and depth can be recovered by matching the bounding box of the generated object to the detected 3D bounding box. This task is typically solved before the reconstruction, at the segmentation and detection phase that happens.

\begin{figure}[ht]
      \centering
        \begin{tabular}{@{}c @{}c @{}c @{}}
            
            \includegraphics[width=.33\linewidth]{images/pix3d/inp1.png} & 
            \includegraphics[width=.33\linewidth]{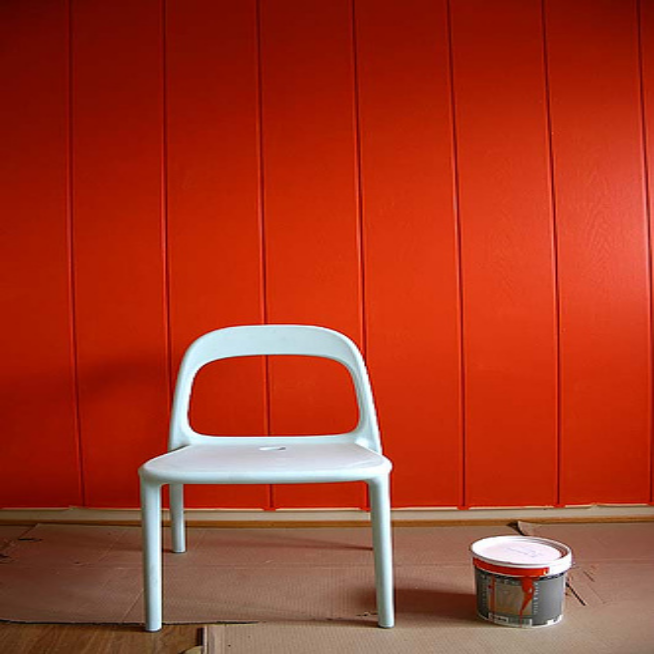} &
            \includegraphics[width=.33\linewidth]{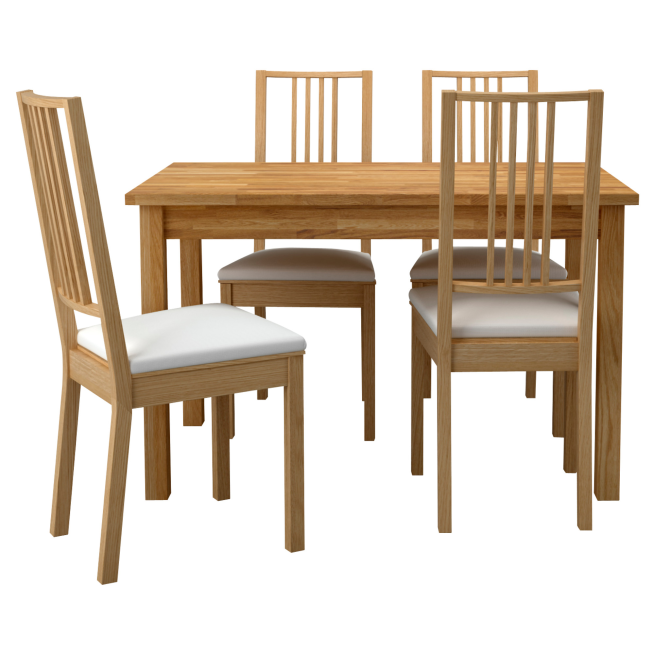}\\
            \multicolumn{3}{c}{Input images.}\\

            
            \includegraphics[width=.33\linewidth]{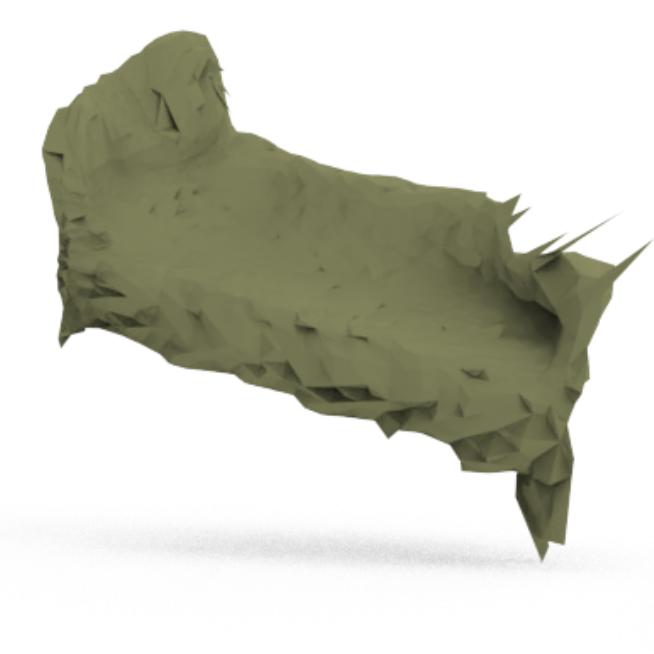} & 
            \includegraphics[width=.33\linewidth]{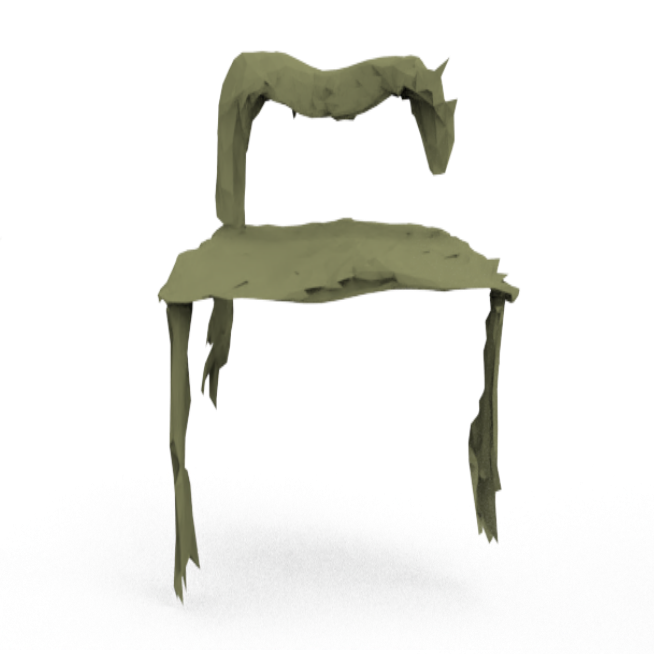}&
            \includegraphics[width=.33\linewidth]{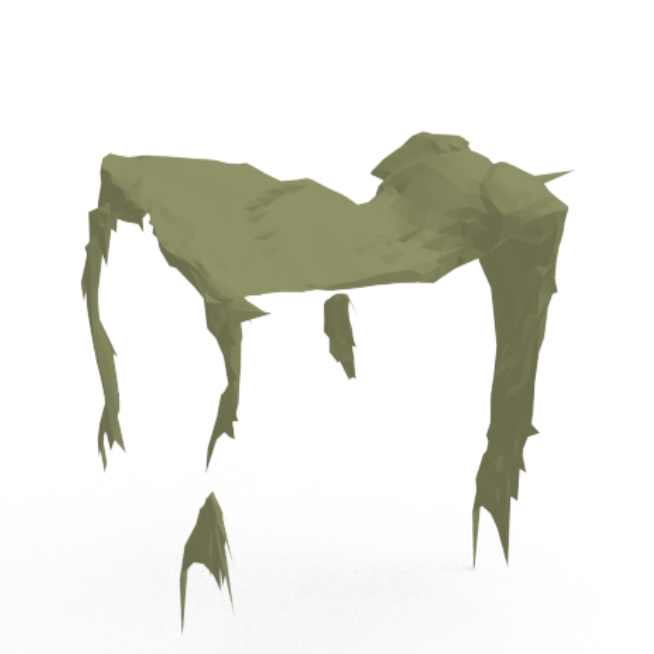}\\
            \multicolumn{3}{c}{3D models reconstructed using Mesh R-CNN.}\\
       
             \includegraphics[width=.33\linewidth]{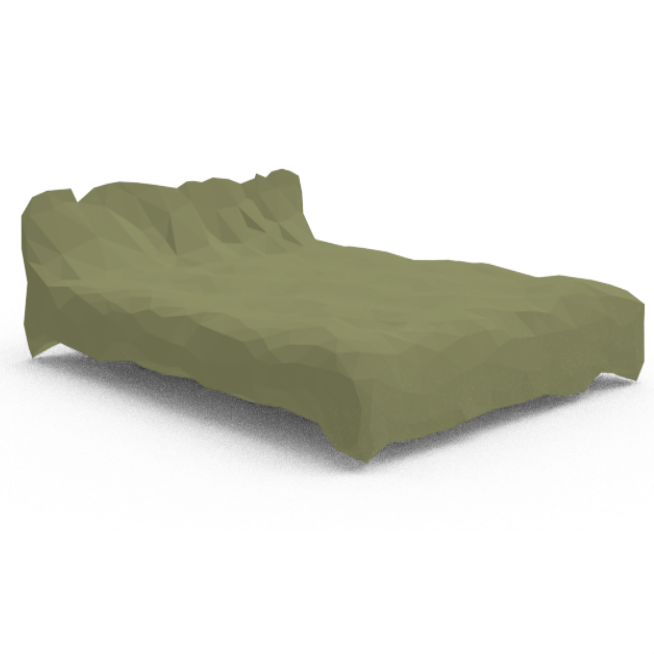} & 
             \includegraphics[width=.33\linewidth]{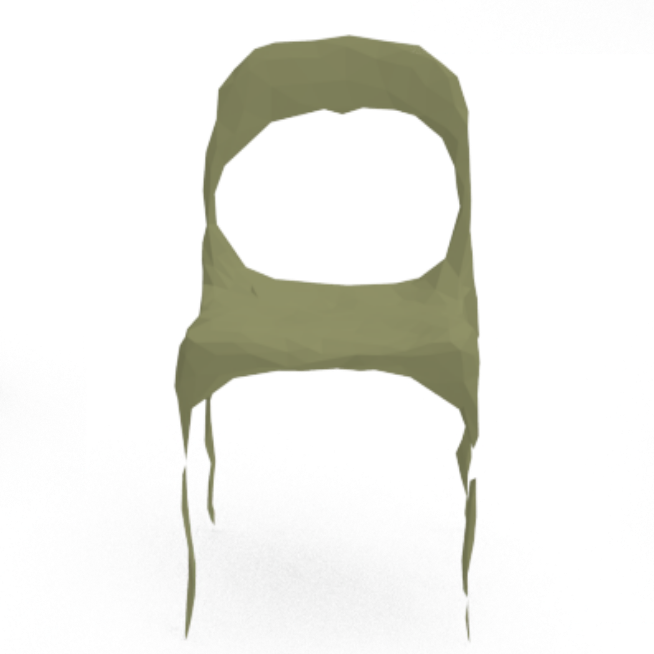}&
             \includegraphics[width=.33\linewidth]{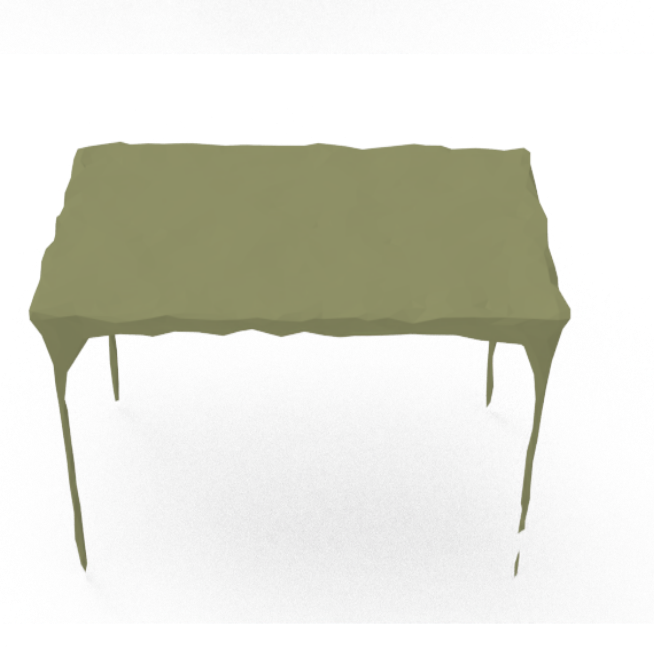} \\
             \multicolumn{3}{c}{3D models reconstructed using the proposed method (Smooth).}
        \end{tabular}
\caption{We compare the visual quality  of the proposed method  (Smooth)  to mesh R-CNN~\cite{gkioxari2019mesh} when trained and tested on real-world images from Pix3d~\cite{sun2018pix3d} benchmark.}
    \label{fig:qual_pix3d}
\end{figure}

Table~\ref{tab:quan_pix3d}  summarises the performance of our model and compares it to Mesh R-CNN. We report a $50\%$ performance increase  in the reconstruction error compared to Mesh R-CNN~\cite{gkioxari2019mesh} 
despite the fact that it does not use the camera intrinsics as is the case with Mesh R-CNN~\cite{gkioxari2019mesh}. Our model also generates visually attractive reconstructions compared to Mesh R-CNN and shown on  Figure \ref{fig:qual_pix3d}.

\begin{table}
\centering
\caption{Quantitative evaluation on Pix3D. We report the CD, the normal consistency, and the F1 scores of Mesh R-CNN and the proposed model (Smooth).  }
\begin{tabular}{@{}l c c c c c@{}}
    \toprule
             \textbf{Model} &  CD$\downarrow$  & Normal$\uparrow$ & F$_{1}^{0.1}\uparrow$ & F$_{1}^{0.3}\uparrow$ & F$_{1}^{0.5}\uparrow$ \\
    \midrule

    Mesh R-CNN & $1.11$ &\textbf{ $0.71$ }& \textbf{$18.7$}&
    \textbf{$56.4$}&
    $73.5$\\

       Ours & \textbf{$0.55$} & $0.73$ & \textbf{$28.3$}&
    \textbf{$70.5$}&
   \textbf{ $83.2$}\\

\bottomrule
\end{tabular}

\label{tab:quan_pix3d}
\end{table}

\subsection{Generalization to in-the-wild images.}
\label{sec:inthewild}
Figure~\ref{fig:wild} evaluates the generalisation capability of the proposed method to in-the-wild images. We take our model, trained on  ShapeNet~\cite{chang2015shapenet}, which is a synthetic dataset, and test it  on real world images including out-of-category images such as luggage (column 1 of Figure~\ref{fig:wild}) and hand bag (column 2 of Figure~\ref{fig:wild}). Our model generalizes well to known categories, since it is able to reconstruct smooth and accurate reconstructions. It also  performs fairly well  on unseen categories. Mesh R-CNN fails to generalise to real world images and it seems that the model has over-fitted ShapeNet~\cite{chang2015shapenet}, which is a synthetic dataset.

\subsection{Ablation study}
\label{sec:ablation_study}
We demonstrate the importance  of each component of the proposed model and validate the choice of its hyper-parameters. 

\vspace{6pt}
\noindent\textbf{Effect of the PPN.} To better quantify the improvement brought by the reinforcement learning, we assess the effect of thresholding values on two shape categories: Chair and Cabinet. While the chair category is composed of objects that contain a large set of complex topological structures and thin parts representing the chair legs and armrests, cabinet category is mostly composed of genus zero objects. Intuitively, a high thresholding value would result in a better accuracy in Cabinet category and low values will achieve good performance for the chair category.  Table~\ref{tab:ab_rl}  compares the reconstruction error of the coarse mesh  when using a fixed triangulation thresholding  compared to the reinforcement learning-based thresholding. A high thresholding value (\eg $0.11$) is, on average, more suitable than a low thresholding value (\eg $0.05$) for the Cabinet category. We observe the opposite behavior on the Chair category, \ie low thresholding values perform better than high ones in terms of F1 scores as seen in Table~\ref{tab:ab_rl}. This reinforces the need for a smart thresholding selection system, which we model using the proposed PPN. In fact, the PPN manages to get the best of the thresholding values including intra-category objects as seen in Table~\ref{tab:ab_rl} where the performance of our Reinforcement learning-based model outperforms every fixed thresholding model on the cabinet category.
\begin{table}
\centering
\caption{Comparative analysis of different thresholding values of Chair and Cabinet categories. We report the $F_ {1}^{r}$, with $r = 1e-4$, of $3000$ surface extracted points on unscaled objects of ShapeNet dataset.}

\begin{tabular}{@{}l @{ } l c c @{}}
    
    \toprule
             &\textbf{Model} &  Cabinet  & Chair   \\
    \midrule

    Fixed      & $\tau= 0.05$ & $30.92$ & \textbf{34.82 } \\
    threshold  & $\tau= 0.085$ & $31.24$ & $31.40$  \\
    $\tau$     &  $\tau= 0.11$ & $32.42$ & $28.43$  \\
    \hline
        & $\tiny{\tau \in \{ 0.05 , 0.085 , 0.11\}}$ & \textbf{33.12} & \textbf{34.82}  \\
\bottomrule
\end{tabular}
    
    \label{tab:ab_rl}
\end{table}

\begin{figure*}[!ht]

\resizebox{0.95\linewidth}{!}{
    \begin{tabular}{ccccc} 
        \includegraphics[width=0.2\linewidth]{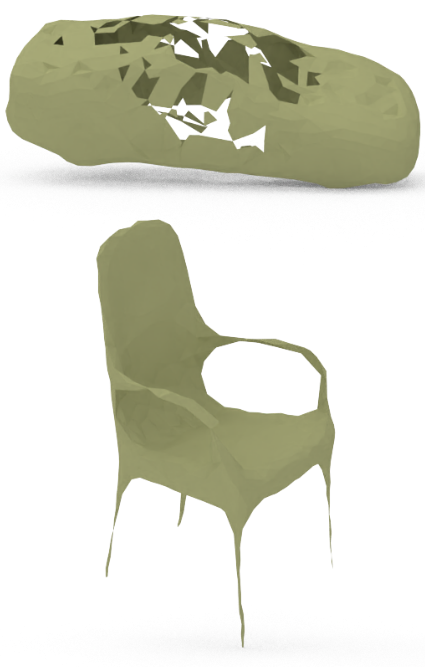}& \includegraphics[width=0.2\linewidth]{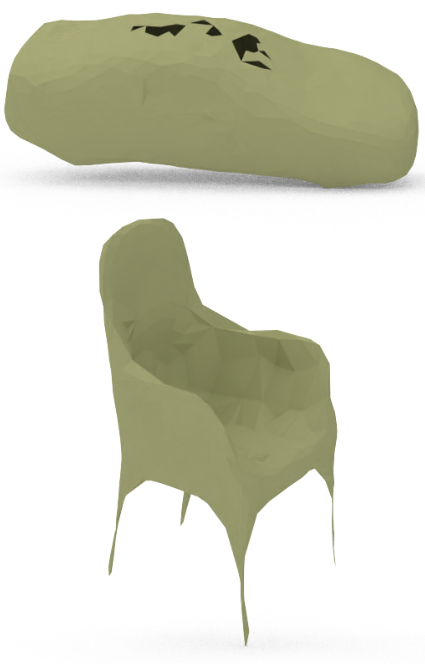}& \includegraphics[width=0.2\linewidth]{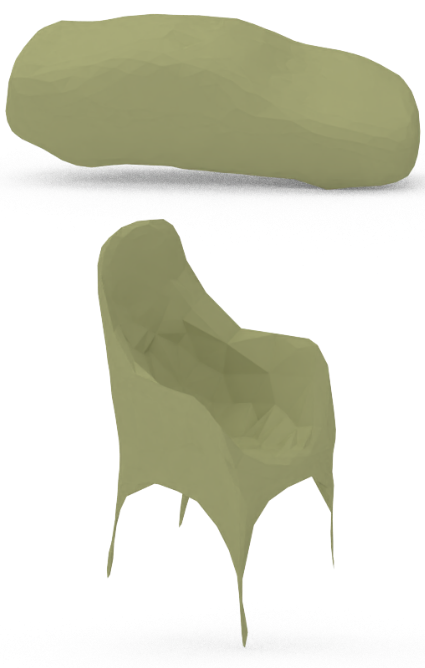}& \includegraphics[width=0.2\linewidth]{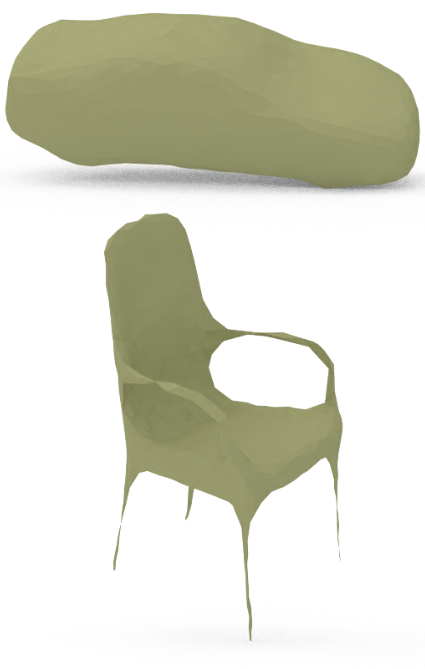}& \includegraphics[width=0.2\linewidth]{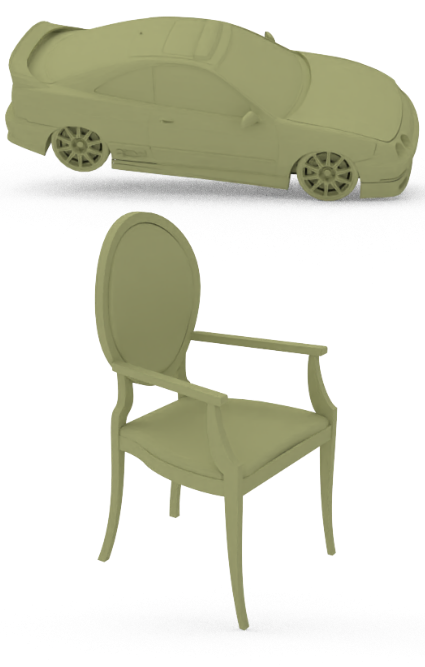}\\
        \small{(a)}   & \small{(b)} & \small{(c) } & \small{(d) Base}  & \small{(e) Ground-}\\
        \small{$\tau=0.05$}   & \small{$\tau=0.085$} & \small{$\tau=0.11$.} & \small{model.}  & \small{truth.}
    \end{tabular}
    }
      \caption{Comparison of different ablated models.  The choice of the threshold $\tau$ affects the quality and  accuracy of the reconstruction. A low threshold $\tau$ is suitable for chairs while a large threshold $\tau$ is more suitable for cars. Our base model was trained using reinforcement learning and is able to select the appropriate threshold that best suits each  input image.} 
      \label{fig:ablation}
      
\end{figure*}

Figure~\ref{fig:ablation} compares the visual aspect of models that use a fixed threshold value $\tau$ and our reinforcement learning-based model (referred to as Base model). A small fixed threshold $\tau=0.05$ results in high fidelity when dealing with objects of complex topological structures, \eg chairs (see Figure~\ref{fig:ablation}-(a)). However, using a small threshold $\tau$ to triangulate genus zero objects such as cars results in  holes (see Figure~\ref{fig:ablation}-(a) and Figure~\ref{fig:ablation}-(b)). A high threshold $\tau$ is more suitable for  genus zero surfaces (see Figure~\ref{fig:ablation}-(c)). Our proposed approach (Figure~\ref{fig:ablation}-(d)) automates the threshold selection process by automatically learning and predict the appropriate threshold value for each object.

\begin{table}
\centering
 \caption{Comparative analysis of different thresholding values. We report the Chamfer distance (in millimeter) evaluated between unscaled models of ShapeNet benchmark and our final reconstructions. A larger model results in overfitting. The lower the values the  better.}
\begin{tabular}{ @{ } l c c @{}}
    
    \toprule
             \textbf{Model} &  Train Set  & Holes test set   \\
    \midrule

         $\tiny{\tau \in \{ 0.05 , 0.085 , 0.11\}}$ & 0.696 &\textbf{ 1.282} \\
         
         $\tiny{\tau \in \{0.035 +i\times\frac{3}{200}\},}$ & \textbf{0.563} & 1.319  \\
         $ i \in \{0, \cdots, 7\}$\\

\bottomrule
\end{tabular}
   \label{tab:ab_th}
\end{table}

\vspace{6pt}
\noindent\textbf{Effect of the number of filtering actions.} The filtering actions needed depends on the shape, topology and size of the target object. Allowing the model to choose from a larger set of thresholds, which results in a larger set of filtering actions, results in overfitting, \ie a lower training error but a higher test error,  as shown in Table~\ref{tab:ab_th}. In this paper, we use three filtering actions. This number  strikes a balance between the model's capability to reconstruct objects and its capability to generalize to unseen objects.

\begin{figure}[ht]

\resizebox{0.95\linewidth}{!}{
    \begin{tabular}{ c c }
       \includegraphics[width=.5\linewidth]{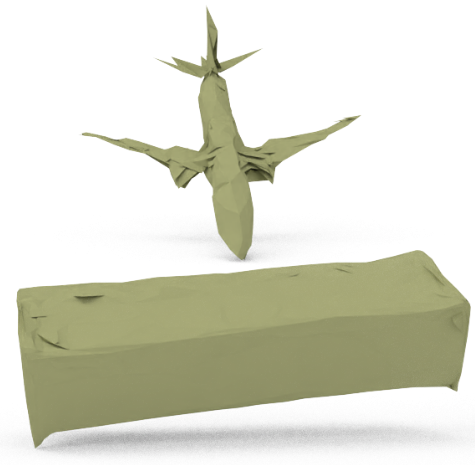}& \includegraphics[width=.5\linewidth]{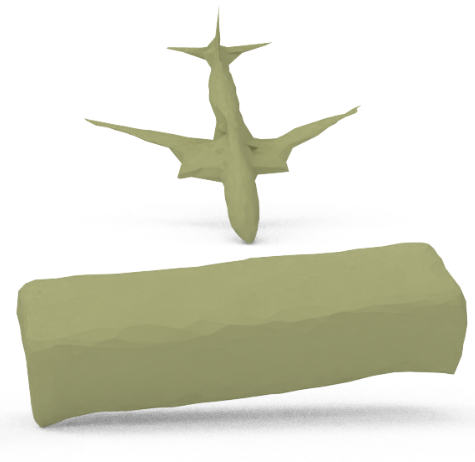} \\
        
        Ours (Pretty) & Ours (Smooth) \\

    \end{tabular}
}

\caption{ Qualitative comparison  between the proposed Pretty and Smooth models.  Ours (Pretty), which was trained following Mesh R-CNN~\cite{gkioxari2019mesh} Pretty protocol, exhibits some topological degeneracy. Ours (Smooth), which  is more regularized, generates smoother surfaces.   }
\label{fig:qual_smooth}
\end{figure}

\vspace{6pt}
\noindent\textbf{Effect of the smoothness term of the loss function.} The smoothness regularization encourages smooth reconstructions and regulates the network to preserve the clean connectivity generated for the coarse mesh. As seen in Figure~\ref{fig:qual_smooth}, smooth reconstructions do not suffer from face self-intersections, unlike Mesg R-CNN (Pretty) where such topological degeneracies can occur; see   Figure~\ref{fig:ResultsShapeNettab} rows 3, 5, and 6. 

\section{Conclusion}
\label{sec:conclusion}
We proposed a novel deep learning-based method for single view-based 3D object reconstruction. In contrast to previous works, our method is able to reconstruct high resolution 3D objects of arbitrary genus without using templates or volumetric grids. The quantitative and qualitative evaluation on the ShapeNet and Pix3D benchmarks  show that the proposed approach outperforms  the current state-of-the-art reconstruction methods. We have also found that the Chamfer distance does not reflect the visual quality of the reconstruction, and thus better measures and better evaluation metrics need to be explored in the future to evaluate the visual quality of the reconstructed meshes but also to guide the training process.   
Despite the performance of the proposed approach, there are several avenues for future research. For instance, our method inherits the limitation of the Chamfer distance which we use as the main criterion to evaluate our prediction. Also, the conversion of meshes into point clouds, which is  crucial to computing the loss,   results in information loss and favours large surfaces over small ones. We plan to address these issues in the future.

\bibliographystyle{IEEEtran}
\bibliography{bib}
\end{document}